\documentclass[acmtog,nonacm,screen]{acmart}
\AtBeginDocument{%
  \providecommand\BibTeX{{%
    \normalfont B\kern-0.5em{\scshape i\kern-0.25em b}\kern-0.8em\TeX}}}

\setcopyright{acmlicensed}
\copyrightyear{2024}
\acmYear{2024}
\acmDOI{10.1145/3680528.3687617}
\acmBooktitle{SIGGRAPH Asia 2024 Conference Papers (SA Conference Papers '24), December 3--6, 2024, Tokyo, Japan}

\acmConference[SA Conference Papers '24]{SIGGRAPH Asia 2024 Conference Papers}{December 3--6, 2024}{Tokyo, Japan}
\acmISBN{979-8-4007-1131-2/24/12}

\acmSubmissionID{509}

\definecolor{DeltaColor}{rgb}{0.039,0.73,0.71}
\definecolor{SetaColor}{rgb}{0.867, 0.0235, 0.376}
\definecolor{SigmaColor}{rgb}{0.98,0.45,0.0}
\definecolor{RedColor}{rgb}{0.8,0,0}
\definecolor{AlphaColor}{rgb}{1.0, 0.4, 0.8}
\definecolor{BetaColor}{rgb}{0.8,0,0.8}
\definecolor{GammaColor}{rgb}{0.0,0,0.7}
\definecolor{EpsilonColor}{rgb}{0.353,0.725,0.906}
\definecolor{TauColor}{rgb}{0.423,0.235,0.192}

{}

\newcommand{\ReprName}{NGC}

\newcommand{\parahead}[1]{\noindent \paragraph{#1}}

\newcommand{\Curve}{\mathcal{C}}
\newcommand{\Point}{\mathbf{p}}
\newcommand{\AnotherPoint}{\mathbf{q}}
\newcommand{\CurveFrame}{\mathcal{F}}
\newcommand{\Feat}{\mathbf{f}}
\newcommand{\Cylinder}{\mathcal{G}}
\newcommand{\GlobCode}{z_{\text{glob}}}
\newcommand{\MLP}{\textit{MLP}}
\newcommand{\NetParam}{\Theta}

\usepackage{amsfonts}

\begin{document}

\title{Controllable Shape Modeling with Neural Generalized Cylinder}

\author{Xiangyu Zhu}
\email{uhzoaix@gmail.com}
\affiliation{%
 \institution{The Chinese University of Hong Kong, Shenzhen}
 \city{Shenzhen}
 \country{China}
}

\author{Zhiqin Chen}
\email{zchen@adobe.com}
\affiliation{%
 \institution{Adobe Research}
 \city{Seattle}
 \country{USA}
}

\author{Ruizhen Hu}
\email{ruizhen.hu@gmail.com}
\affiliation{%
 \institution{Shenzhen University}
 \city{Shenzhen}
 \country{China}
}

\author{Xiaoguang Han}
\email{hanxiaoguang@cuhk.edu.cn}
\affiliation{%
 \institution{The Chinese University of Hong Kong, Shenzhen}
 \city{Shenzhen}
 \country{China}
}
\authornote{Corresponding author: hanxiaoguang@cuhk.edu.cn}

\renewcommand{\shortauthors}{Zhu, et al.}

\begin{teaserfigure}
\centering
\includegraphics[width=1.0\linewidth]{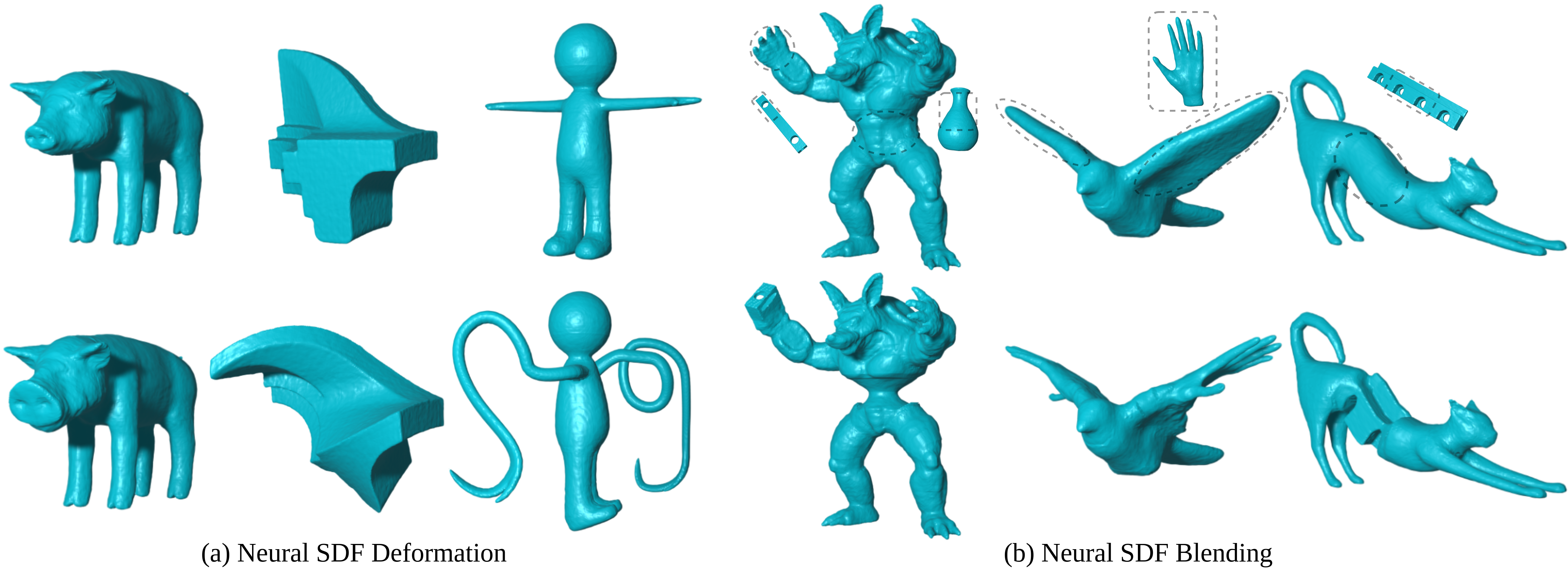}
\caption{Shape modeling with our proposed neural generalized cylinder (\ReprName) representation.
(a) After fitting a shape with \ReprName, we can deform the shape by manipulating a few control points defined on generalized cylinders to achieve local scaling, twisting, and complex curved deformations (the stickman at third column poses like the word \textit{sig}).
(b) Furthermore, by interpolating neural features learned by \ReprName, distinct shapes can be blended together.
}
    \label{fig:0_teaser}
\end{teaserfigure}

\begin{abstract}

Neural shape representation, such as neural signed distance field (NSDF), becomes more and more popular in shape modeling as its ability to deal with complex topology and arbitrary resolution. Due to the implicit manner to use features for shape representation, manipulating the shapes faces inherent challenge of inconvenience, since the feature cannot be intuitively edited. In this work, we propose neural generalized cylinder (\ReprName)~ for explicit manipulation of NSDF, which is an extension of traditional generalized cylinder (GC). Specifically, we define a central curve first and assign neural features along the curve to represent the profiles. Then NSDF is defined on the relative coordinates of a specialized GC with oval-shaped profiles. By using the relative coordinates, NSDF can be explicitly controlled via manipulation of the GC. To this end, we apply \ReprName~to many non-rigid deformation tasks like complex curved deformation, local scaling and twisting for shapes. The comparison on shape deformation with other methods proves the effectiveness and efficiency of \ReprName. Furthermore, \ReprName~ could utilize the neural feature for shape blending by a simple neural feature interpolation. 


\end{abstract}

%
%

%
%



\maketitle

\section{Introduction}

A convenient way of modeling complex shapes is always pursued in computer graphics. Traditional methods~\cite{ARAP_modeling:2007, funkhouser2004modeling, deformation_IRBF} can edit current shapes by deformation or blending to create new shapes, but it is restricted by 
the quality of mesh discretization, the difficulty of topological change and the robustness of optimization. To overcome these challenges, neural networks were introduced into 3D shape modeling, which brings great successes in tasks like shape reconstruction from images~\cite{wen20223d} and shape generation conditioned on images or texts~\cite{3DShape2VecSet}. Even for common geometry processing, like deformation, NFGP~\cite{yang2021NFGP} has also proved that the superiority of neural representation (neural SDF) than traditional mesh.

Generally speaking, neural networks enable us representing complex shapes as high dimensional feature vectors, and large-scale and type-variant shape modeling tasks can be done by manipulation on these features. 
%
However, such feature manipulation is not intuitive for users to control, as users are used to giving explicit constraints on shapes, such as manipulating some sparse point handles ~\cite{ARAP_modeling:2007}. Thus, an interesting question arises: \emph{Can we also provide explicit handles to neural shape representations?}

Before answering this question, let's review a common-used shape representation - generalized cylinder (GC), which is widely adopted in shape modeling~\cite{Zheng2011Component-wise-Controllers, PAM2014} and interactive surface reconstruction~\cite{arterial_snakes2010, Chen2013_3sweep, Morfit2014}. The core of this representation is a central curve together with cross-sectional profiles, making it a simple and flexible way of deformation control. Inspired by this, our key idea is to introduce neural ways into this explicit representation and define features over the curve to implicitly represent those profiles. In this manner, the skeleton curve can naturally be used as the control handles.

To this end, we propose neural generalized cylinder (\ReprName) for controllable shape modeling. More specifically, we define the neural SDF on the relative coordinates in a specialized \textit{generalized cylinder} (GC) with oval-shaped profiles. 
Intuitively speaking, by using the relative coordinates, we expect the shape of neural SDF can relatively deform with respect to the change of the GC, which is similar to the effect of the relative coordinates used in cage-based deformation methods~\cite{joshi2007harmonic,ju2023mean,lipman2008green,jacobson2011bounded}. 
We assign local features to the central curve, which can be viewed as generalizing the explicit profile to the feature vector, as shown in Fig.~\ref{fig:3_method_vs_GC}.
In classic GC, two shapes can be blended by interpolating their profiles. 
As an analogy, we can interpolate the local features on the central curves from two \ReprName~ shapes, and obtain a blended shape. Some results can be found in Fig.~\ref{fig:0_teaser}. 
\ReprName~ shows superior representational ability than classic GC in the flexibility of the skeleton definition and representing complex profiles.

%


To obtain the \ReprName~representation for a given shape with a set of GCs, we adopt the auto decoder framework as in DeepSDF~\cite{deepsdf}, where each GC has a learnable global feature vector. 
Unlike other neural implicit methods, our network takes the \textit{relative coordinates} of a query point as one of the inputs and predicts the SDF value at the query point. 
Besides the relative coordinates, we need to further define the feature of the query point. The feature of a curve point (a point on the central curve of the GC) can be extracted by a local feature module from the global feature. 
Then for any point in the GC volume, its local feature can be specified by finding the corresponding curve point, which is the center of the profile containing this point. Finally, a multi-layer perceptron (MLP) takes the relative coordinates and local feature of a query point as the inputs, then predicts the SDF value at the query position. 

In experiments, we collect a set of 50 shapes to test the representation capability of  \ReprName~ and show how it can be used for shape modeling.
We compare the fitting results with DeepSDF~\cite{deepsdf}, which proves the capacity of \ReprName~ for representing general shapes. 
For deformation of general shapes, we compare our method with traditional methods~\cite{ARAP_modeling:2007,SR_ARAP} and neural method NFGP~\cite{yang2021NFGP}. 
The results show the effectiveness and efficiency of \ReprName. 
We also show examples of shape blending by simply blending their local features, where parts from distinct shapes can be mixed smoothly. 

To summarize, our key contributions includes:
\begin{itemize}
    \item To the best of our knowledge, we are the first to solve the explicit manipulation of neural SDF with generalized cylinders, and firmly believe this will inspire relevant researches. 
    \item We introduce \ReprName, a novel shape representation, by defining neural SDF on the relative coordinate of generalized cylinders; 
    \item We present modeling results based on \ReprName, including shape deformation and blending of multiple shapes. 
\end{itemize}
\begin{figure*}
    \centering
    \includegraphics[width=\linewidth]{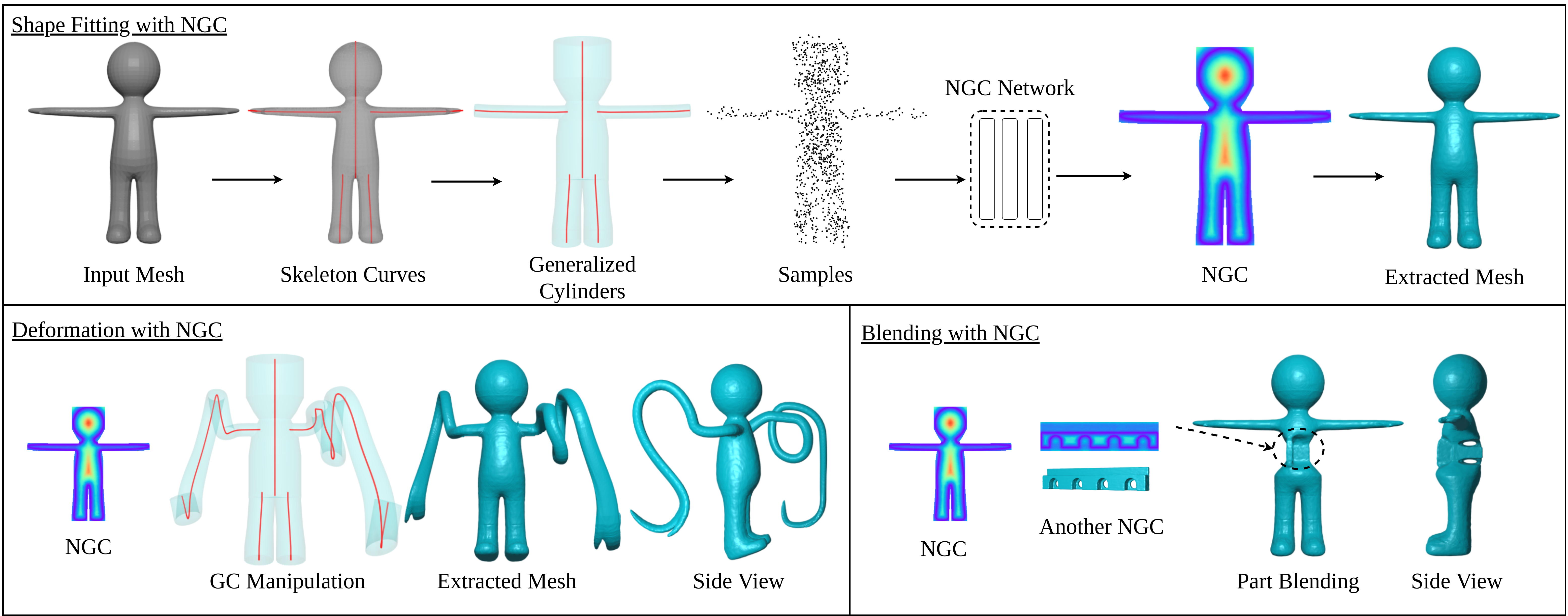}
    \caption{Overview of \ReprName~ for shape fitting, deformation, and blending.
    To convert a input mesh to \ReprName~ representation, we first prepare the skeleton curves and generate corresponding generalized cylinders (GCs). Then we sample points and their signed distances within the GCs to train a neural SDF network, where the neural SDF is defined in the relative coordinate systems of the GCs. Finally, the reconstructed mesh can be extracted by Marching Cubes.
    For deformation, because the neural SDF is defined with relative coordinates on the GCs, we can directly manipulate the GCs in order to deform the shape.
    For blending, after training our NGC network on a collection of shapes, we can blend the part from one shape into another shape by simply replacing and interpolating their learned neural features.
    }
    \label{fig:3_method_overview}
\end{figure*}

\section{Related Works}
In addition to the generalized cylinder, Our method is also closely related to classic and deep learning methods for shape modeling, specifically, shape deformation and shape editing techniques.

\parahead{Classic deformation methods}
The majority of classic shape deformation methods are designed to operate on polygon meshes.
Energy-based optimization methods~\cite{alexa2000rigid,sorkine2007rigid,terzopoulos1987elastically,lipman2004differential,sorkine2004laplacian,zhou2005large,sumner2007embedded,botsch2007linear} deform meshes with respect to keypoints or handles before and after deformation, by optimizing energy functions defined on the surface or volume of the shape to satisfy certain constraints, such as elasticity, smoothness, and rigidity.
Cage-based deformation methods~\cite{sederberg1986free,joshi2007harmonic,ju2023mean,lipman2008green,jacobson2011bounded} enclose the shape by a coarse
cage mesh where the coordinates of mesh vertices are represented as weighted sums of the coordinates of cage vertices, e.g., via generalized barycentric coordinates, so that the deformation of the shape can be controlled by the deformation of the coarse cage mesh.
Similarly, for character animation, skeleton-driven skinning methods~\cite{baran2007automatic,yoshizawa2007skeleton,yoshizawa2003free,aujay2007harmonic,kho2005sketching} bind the mesh vertices to the skeleton joints, therefore the deformation of the shape can be controlled by the movement of the skeleton. 
However, cage-based methods usually need larger number of control points for finer control of shapes due to the nature of discrete cage, which will be tedious works for users. In this work, we opt to establish relative coordinate system on continous curves to achieve control like skeleton-based methods, which could be easier for users to manipulate than cage-based methods.
%
\parahead{Generalized cylinder}
The generalized cylinder (GC) is known for representing the shape or its part as one central curve and some cross-sectional profiles at key locations on the curve. It is easier and more flexible to control due to its minimal control points and parameters, compared to cage-based deformation. Prior works~\cite{gcd15} explore GC decomposition of complex shapes by proposing the cylindricity measure and solving an exact cover problem with over-complete covers of GCs.
GC is frequently adopted for shape modeling~\cite{Zheng2011Component-wise-Controllers} and surface reconstruction~\cite{arterial_snakes2010, Chen2013_3sweep, Morfit2014}.
While GCs can also serve as handles to control mesh deformation~\cite{ma2018skeleton}, we consider adopting GCs for controllable shape modeling with neural implicit representation.

\parahead{Deformation with neural networks}
With the advancement of deep learning techniques, classic mesh deformation methods are combined and improved by neural networks. Deep learning models can predict surface energy functions~\cite{aigerman2022neural}, automatically construct and deform 3D keypoints and coarse cages for cage deformation~\cite{jakab2021keypointdeformer,yifan2020neural}, as well as rigging skeletons for character animation~\cite{xu2020rignet}.
Some other works adopt a deformation field that is parameterized by a neural network, to achieve mesh deformation for 3D shape reconstruction~\cite{groueix2018papier,wang20193dn} and manipulation~\cite{tang2022neural}.
More recent methods adopt the neural implicit representation~\cite{deepsdf, occnet, IMnet}, such as neural Signed Distance Field (SDF), which represents the SDF of a 3D shape with a continuous function parameterized by a neural network.
The neural SDF representation effectively addressed many issues of the mesh representation, such as sensitivity to triangulation and discretization, and inability to perform topological changes.
Deforming the neural field representation is not straightforward, as certain classic algorithms on meshes do not directly apply to fields.
Some works explored cage deformation~\cite{peng2022cagenerf}, which is limited by the discrete cage as previously discussed, and skeleton rigging~\cite{chen2021snarf}, which mainly focuses on human objects, while most works~\cite{park2021nerfies,pumarola2021d} adopt deformation fields.
Energy-based optimization methods were proposed for classic implicit fields~\cite{museth2002level,osher2004level}, albeit with discretized voxels or octrees as shape representations.
NFGP~\cite{yang2021NFGP} propose to directly optimize the neural SDF without such discretization, to perform shape deformation and other operations.
In comparison, we propose a new representation that allows easy control and deformation of the neural SDF, and shows superior flexibility, shape quality, and speed.

\parahead{Shape editing}
In addition to shape deformation, there are various other shape editing operations, such as object insertion, part replacement, and style transfer~\cite{haque2023instruct}. In this work, we mainly consider part replacement, or part mixing~\cite{funkhouser2004modeling}. Prior works have explored part-based modeling and manipulation~\cite{li2021sp,yin2020coalesce,hertz2022spaghetti} with neural networks, yet they are designed to work on specific shape categories that are determined by their dedicated training set.
Our method is designed to work on general shapes, with the additional capability of part-level mixing or interpolation.

\section{Method}

In this section, we first define the relative coordinate system in a generalized cylinder (GC) in Sec.~\ref{sec:method_coord}, which is crucial in our subsequent derivation of neural generalized cylinders (\ReprName) in Sec.~\ref{sec:method_neural}, where we approximate a shape in the relative coordinate system with neural SDF. We introduce several applications enabled by our representation, including shape deformation and shape blending. An overview of our method is shown in Fig.~\ref{fig:3_method_overview}.

\subsection{Relative Coordinate System}
\label{sec:method_coord}

\begin{figure}
    \centering
    \includegraphics[width=\linewidth]{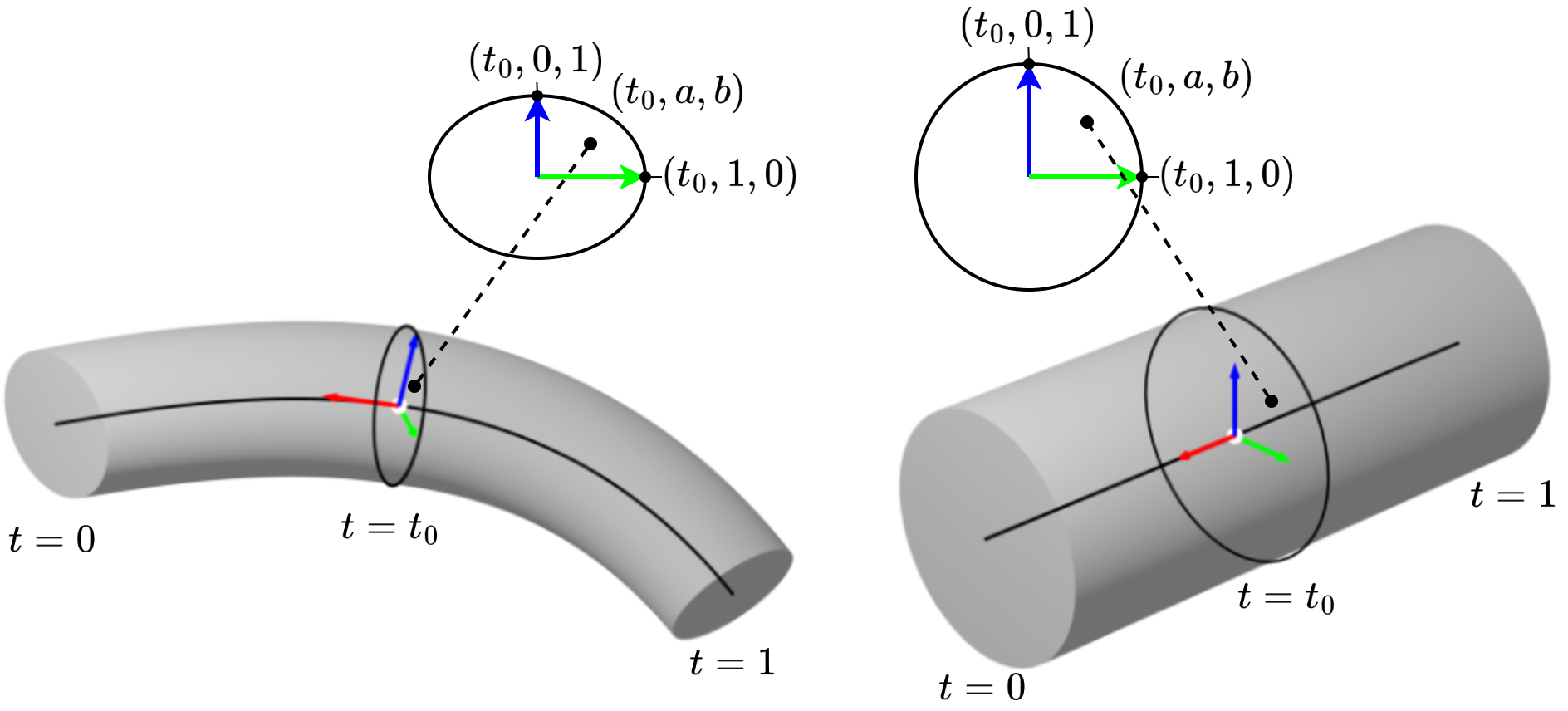}
    \caption{Illustration of the relative coordinate system used in our method. Red, green, and blue indicate the x, y, and z axes, respectively. The position of a point can be specified by its relative coordinates, which are invariant to the ``shape'' of the GC.}
    \label{fig:3_method_relative_coord}
\end{figure}

Our relative coordinate system is defined inside a GC,  as shown in Fig.~\ref{fig:3_method_relative_coord}. Therefore, we first formally define the GC used in our case, and then introduce the relative coordinate system.
Notice that the GC used in our method is a special case of GC. For convenience, we refer to GC in the general shape modeling literature as \textbf{classic GC}, and those specialized GC in our method as \textbf{GC}.

\parahead{Definition of GC}
The axis of our GC is a 3D curve defined by a continuously differentiable function $\Curve$ such that the points on the curve are defined by $(x(t),y(t),z(t)) = \Curve(t), t \in [0,1]$. For each point $\Point = \Curve(t)$ on the curve, we define an orthogonal coordinate system for the local frame, where the origin is at $\Point$, the x axis is the tangent vector of the curve on $\Point$, i.e., $\mathbf{x}(t)=\Curve^{\prime}(t)$, and the y and z axes are defined by some given continuous functions $\mathbf{y}(t)$ and $\mathbf{z}(t)$; see Fig.~\ref{fig:3_method_relative_coord} for examples. Notice that the cross-sectional profile\footnote{We consider not only the boundary curve, but also the area of the shape in the cross-section as the profile.} of the GC at $t$ can be defined as a 2D shape in the yz-plane of $\Point$. An oval-shaped profile can be defined as:
\begin{equation}
\textit{Profile}(t) = 
\{\AnotherPoint \; | \; \CurveFrame^t_x(\AnotherPoint) = 0 \;\textrm{and}\; \CurveFrame^t_y(\AnotherPoint)^2 + \CurveFrame^t_z(\AnotherPoint)^2 \le 1 \},
\end{equation}
where $\CurveFrame^t$ converts the coordinates from the world frame (the global reference frame in $\mathbb{R}^3$) to $\Point$'s local frame, i.e., $\CurveFrame^t_x(\AnotherPoint) = (\AnotherPoint-\Point) \cdot \mathbf{x}(t)$, $\CurveFrame^t_y(\AnotherPoint) = (\AnotherPoint-\Point) \cdot \mathbf{y}(t)$, and $\CurveFrame^t_z(\AnotherPoint) = (\AnotherPoint-\Point) \cdot \mathbf{z}(t)$. Also note that the longest and shortest radii of the oval are implicitly defined by the length of $\mathbf{y}(t)$ and $\mathbf{z}(t)$, i.e., $r_y(t) = \frac{1}{||\mathbf{y}(t)||}$ and $r_z(t) = \frac{1}{||\mathbf{z}(t)||}$. Therefore, the shape of our GC is determined by three functions $\Curve, \mathbf{y}, \mathbf{z}$; and defined as the union of all oval-shaped profiles for all points on the curve:
\begin{equation}
\Cylinder(\Curve, \mathbf{y}, \mathbf{z}) = \bigcup_{t \in [0,1]} \textit{Profile}(t).
\end{equation}

\parahead{Relative coordinate system at GC}
Given the definition of the GC, it is easy to find a relative coordinate system for points inside the GC that is invariant with respect to the ``shape'' of the GC. Specifically, we consider the relative coordinates $(t, a, b)$, such that the world coordinates of the point is $\AnotherPoint = (\CurveFrame^t)^{-1}([0, a, b])$, where $(\CurveFrame^t)^{-1}$ converts the input point from the local frame with respect to $\Curve(t)$ into the world frame. Intuitively, the relative coordinates specify that the point is located on the profile of point $\Curve(t)$, and its location on the profile (yz-plane of $\Curve(t)$'s local frame) is $(a, b)$, as shown in Fig.~\ref{fig:3_method_relative_coord}. Notice that the world coordinates of $\AnotherPoint$ naturally satisfies $\Curve^{\prime}(t) \bot (\AnotherPoint - \Curve(t))$.

\begin{figure}
    \centering
    \includegraphics[width=\linewidth]{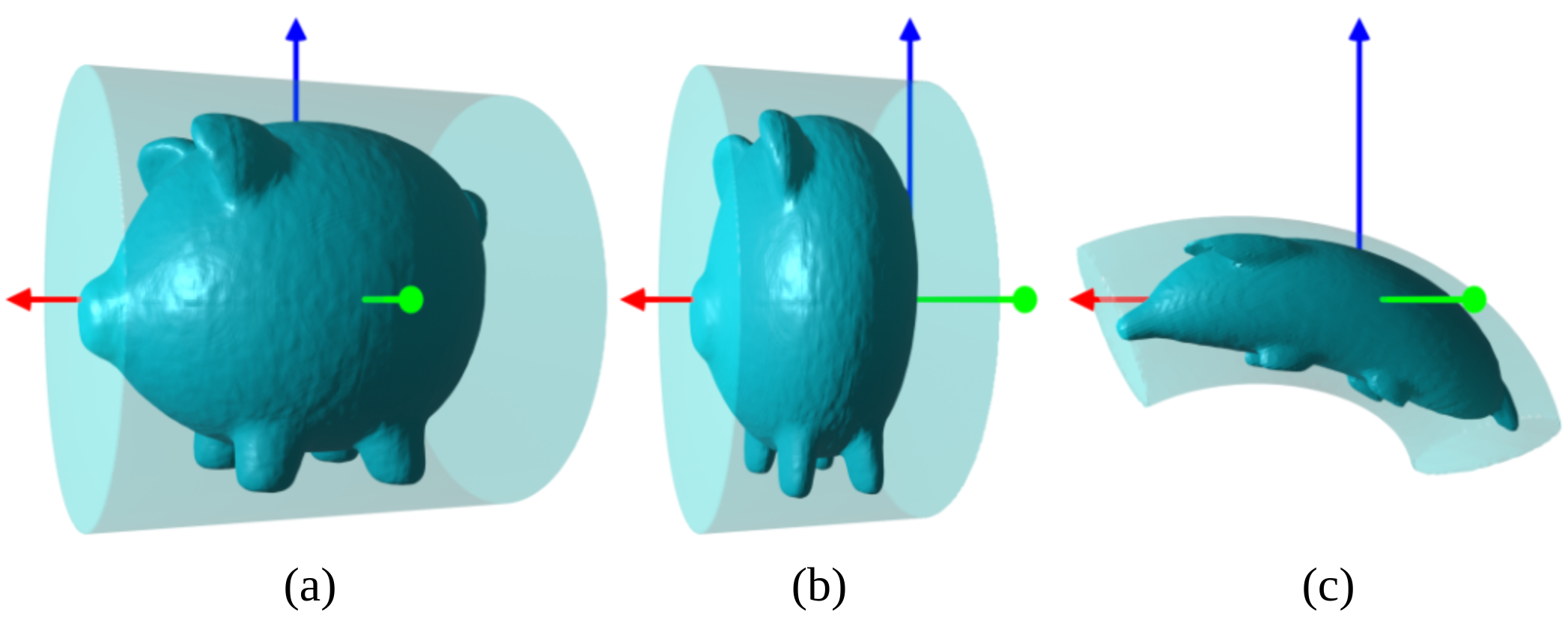}
    \caption{(a) An implicit field (visualized with its 0-isosurface) is defined on the relative coordinates of a GC. (b) Canonical cylinder. (c) The implicit field will change according to changes of the GC. 
    }
    \label{fig:3_method_canonical_cylinder}
\end{figure}

\parahead{Properties of the relative coordinates}
Since our coordinate system is relative to $\Curve$, $\mathbf{y}$, and $\mathbf{z}$, which fully define the shape of a GC, one can expect that an implicit field defined on our relative coordinate system will naturally deform with the GC when the GC changes its shape, so that the shape of the implicit field stays the same with respect to the shape of the GC. An example is shown in Fig.~\ref{fig:3_method_canonical_cylinder}. Note that when the shape of the GC is a canonical cylinder ($[0,1] \times [-1,1]^2$) as shown in Fig.~\ref{fig:3_method_canonical_cylinder} (b), the world coordinates $(x,y,z)$ are equal to the relative coordinates $(t,a,b)$. The implicit field defined in this GC can be considered as having a canonical pose. When the GC changes shape, as in Fig.~\ref{fig:3_method_canonical_cylinder} (c), the implicit field will deform accordingly.

\parahead{Mapping from world coordinates to relative coordinates}
Note that there may be multiple relative coordinates that map to the same world coordinates, because a point $\AnotherPoint$ may be located on profiles at multiple $t$'s, i.e., the equation $\Curve^{\prime}(t) \cdot (\AnotherPoint - \Curve(t)) = 0$ can have multiple solutions for $t$. Therefore it can bring challenges in computing the relative coordinates from world coordinates. To address the ambiguity, for a point $\AnotherPoint$ in the world coordinates, we first find its closest point on the curve, so that $t = \arg \min_t || \AnotherPoint - \Curve(t) ||_2$. When $t$ is determined, we can easily compute $a = \CurveFrame^t_y(\AnotherPoint)$ and $b = \CurveFrame^t_z(\AnotherPoint)$. Note that when $t\neq\{0,1\}$, $\Curve^{\prime}(t) \bot (\AnotherPoint - \Curve(t))$ due to the curve being continuously differentiable, therefore $\CurveFrame^t_x(\AnotherPoint)=0$, so that $\AnotherPoint$ is guaranteed to be on the profile of the GC at point $\Curve(t)$, which shows that the computed $(t,a,b)$ is indeed one of the solutions of the world-to-relative mapping. Some edge cases will be discussed in the supplementary material.

\parahead{Implementation details}
In our experiments, we use Bezier curve to represent each curve $\Curve$. To facilitate fast closest-point computation, the curve is discretized into a piece-wise linear curve before computing relative coordinates. To define $\mathbf{y}$ and $\mathbf{z}$, we identify several key frames on the curve, e.g., $t=\{0.0, 0.5, 1.0\}$, and define $\mathbf{y}(t)$ and $\mathbf{z}(t)$ at those locations. For the rest of $t \in [0,1]$, $\mathbf{y}(t)$ and $\mathbf{z}(t)$ can be obtained via a combination of linear interpolation and spherical linear interpolation (Slerp) with those defined values; see supplementary material for more details.

At this point, we have defined the GC, the relative coordinate system, and the mappings between world frame and relative frame. Now we are ready to introduce Neural Generalized Cylinder, where a 3D shape is represented by a neural SDF operating on the relative coordinate system in a GC.

\subsection{Neural Generalized Cylinder}
\label{sec:method_neural}
\begin{figure}
    \centering
    \includegraphics[width=\linewidth]{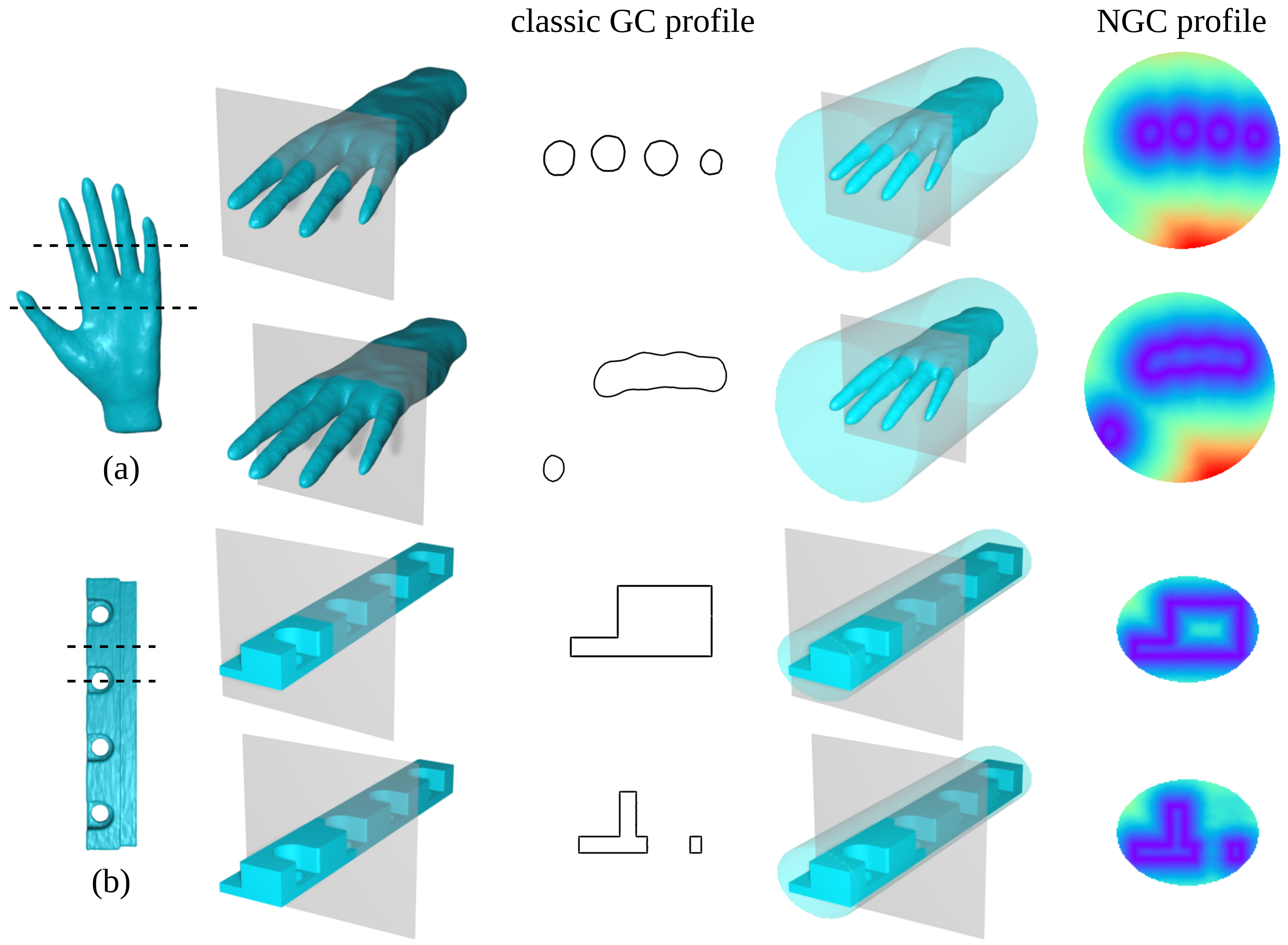}
    \caption{
    Visualization of profiles for classic GC and \ReprName. Classic GC stores a limited number of profiles for each shape. The profiles between two existing profiles are obtained via profile interpolation, which is challenging when the two profiles are with different topology. The topological difference between profiles could be caused by non-convex shape as in (a) or high genus as in (b). In contrast, \ReprName~ represents all profiles as continuous neural SDF, thereby avoiding the interpolation challenge.
    }
    \label{fig:3_method_vs_GC}
\end{figure}

\parahead{Comparison with classic GC. }
In GCD~\cite{gcd15}, a 3D shape is decomposed into multiple parts, where each part is a \textit{classic GC} represented as a 3D curve and several 2D profiles at key frames on the curve.
Fig.~\ref{fig:3_method_vs_GC} left shows two example parts, and their cross-sectional profiles if they were to be represented as classic GCs.
Typically, to obtain the complete shape of each part, the profile at any point on the central curve can only be interpolated from adjacent profiles defined at key frames.
However, it is a significant challenge to interpolate profiles with different topology, as in Fig.~\ref{fig:3_method_vs_GC} left.
Therefore, classic GCs have considerable difficulty fitting general shapes, while in contrast, our NGC approach does not have such issues.
In our method, we generalize the classic GC into NGC, where each part is fully enclosed by a GC as defined in Sec.~\ref{sec:method_coord}, and the shape of the part is represented by a neural SDF defined in the relative coordinate system of the GC.
As shown in Fig.~\ref{fig:3_method_vs_GC} right, since our neural SDF is defined everywhere inside the GC, we are able to extract the shape without interpolating profiles, demonstrating the superior representational ability of NGC compared to classic GCs.
In the following, we will first describe how neural networks can be applied to fit a single part with a single GC. Then we will introduce training our network on a collection of shapes and parts. After training, we can perform mesh extraction and shape editing.

\parahead{Preprocessing}
Given the input triangle mesh, we aim to cover the mesh with a set of GCs to establish the relative coordinate systems before training our \ReprName. As shown in Fig.~\ref{fig:3_method_overview}, we first extract the skeleton of the shape, which consists of several curves and their key points. Afterwards, the radii and local frames at the key points on the GC can be easily estimated, so that all parameters of the GC are obtained and the shape of the GC is determined.

\begin{figure}
    \centering
    \includegraphics[width=\linewidth]{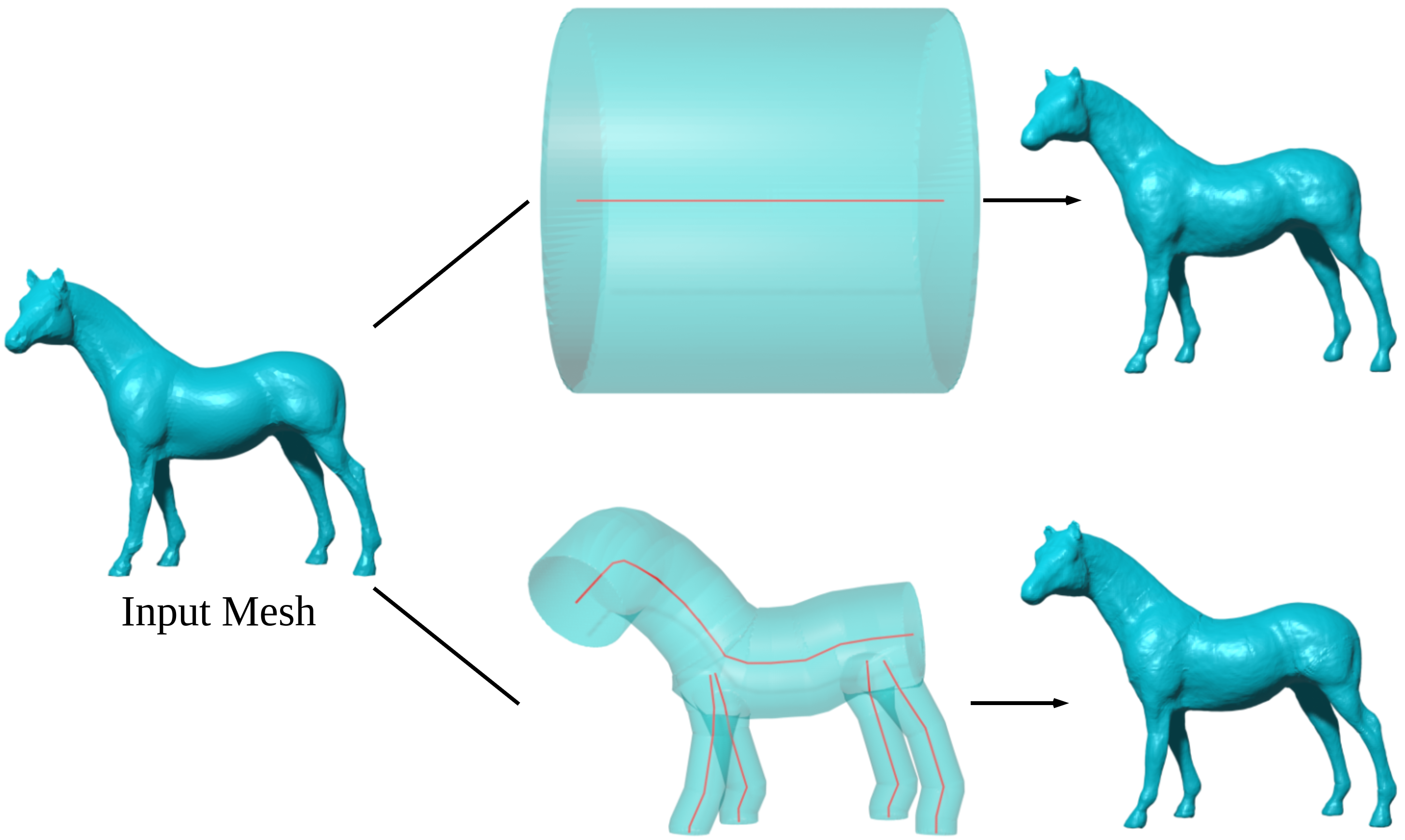}
    \caption{
    The flexibility of skeleton definition in \ReprName. Bottom: we can extract the skeleton curves and generate a compact GC covering, and represent the shape with \ReprName. Top: alternatively, we can use just one large GC (with one curve) to represent the shape. The two reconstructed meshes are similar in visual quality, but the difference is in controllability. More complex control of the shape is supported by the finer GCs. 
    }
    \label{fig:3_method_skeleton_flexible}
\end{figure}

Meso skeleton~\cite{meso_skeleton} can be used for skeleton extraction of general shapes, but it is sensitive to geometric details of the surface and hence may not produce a simple curve skeleton for a shape with complex topology, e.g., Fig.~\ref{fig:3_method_vs_GC} (b), and its predicted skeleton can be found in Fig.~\ref{fig:supp_meso_skeleton}.
Generalized Cylinder Decomposition (GCD)~\cite{gcd15} is more suitable for curve extraction in certain cases, but it struggles when handling shapes with intense variation of profiles. On the other hand, unlike GCD, \ReprName~ allows a more flexible definition for the skeleton of a shape as shown in Fig.~\ref{fig:3_method_skeleton_flexible}, due to our neural SDF representation of shapes. Therefore, the skeletons can be defined based on how users want to control the shapes. In the end, we opt to annotate the curves manually for shapes used in our experiments. Detailed skeletons of all data can be found in the supplementary material.

\parahead{Network design}
A classic GC is made of a central curve and profiles at points on the curve. Therefore, we naturally define a local feature vector $\Feat(t) \in \mathbb{R}^{512}$ for each point $\Curve(t)$ on the curve, $t \in [0,1]$, so that the feature effectively represents the profile at that point. In practice, we define one global feature vector $\GlobCode \in \mathbb{R}^{512}$, and obtain the feature vectors at the endpoints of the curve via $\Feat(0) = \MLP_0(\GlobCode)$ and $\Feat(1) = \MLP_1(\GlobCode)$, where $\MLP_0$ and $\MLP_1$ are learnable multi-layer perceptrons (MLPs). The feature for the rest of $t$ can be interpolated by $\Feat(t) = t \Feat(0) + (1-t) \Feat(1)$.
Therefore, the neural SDF $F$ defined on the relative coordinates $\Bar{\Point} = (t,a,b)$ is
\begin{equation}
F(\Bar{\Point}; z) = h(t,a,b, g(\Feat(t)) ),
\end{equation}
where $g$ and $h$ are MLPs. Recall that to find the relative coordinates of a point in world coordinates, $t$ is determined by its closest point on the central curve. Therefore, the predicted signed distance of a point depends on the feature of its closest point $\Point$ on the curve, and its relative coordinates on the profile at $\Point$. Detailed MLP architectures can be found in supplementary material.

\parahead{Shape fitting}
To fit our \ReprName~ on a set of 3D objects, we assume that each object has already been approximated by a number of GCs, where the union of all the GCs fully enclose the object, as discussed in Sec.~\ref{sec:method_neural}. Assume we have $N_{\Cylinder}$ GCs in total. 
For each GC, we sample 3D points within the volume of the GC and compute their signed distances with respect to the object.
We denote each GC as $\Cylinder_i$; denote the sampled points in $\Cylinder_i$ in relative coordinates as $\Bar{\Point} \sim \Cylinder_i$, and the signed distance of $\Bar{\Point}$ as $\textit{sdf}_i(\Bar{\Point})$.
We adopt the auto-decoder proposed in DeepSDF~\cite{deepsdf}, so that the global feature vector $z_i$ for $\Cylinder_i$ is optimizable. 
Therefore, the training objective is
\begin{equation}
\min_{\forall z_i, \NetParam} \mathcal{L}_{data}  + \lambda \mathcal{L}_{reg}
\end{equation}
where $\NetParam$ is the parameters of all MLPs in \ReprName, including $\MLP_0$, $\MLP_1$, $g, h$. $\mathcal{L}_{data},\mathcal{L}_{reg}$ can be defined as:
\begin{align}
\mathcal{L}_{data} & = \frac{1}{N_{\Cylinder}} \sum_{i=1}^{N_{\Cylinder}} \mathbb{E}_{\Bar{\Point} \sim \Cylinder_i} \left [ \left | F(\Bar{\Point}; z_i) - \textit{sdf}_i(\Bar{\Point}) \right | \right ] \\ 
\mathcal{L}_{reg} & = \frac{1}{N_{\Cylinder}} \sum_{i=1}^{N_{\Cylinder}} ||z_i||_2^2.
\end{align}

We set $\lambda=0.0001$ in the experiments. We also follow DeepSDF's training schedule, to optimize only $z_i$ while freezing the other parameters in the second training phase.

\parahead{Mesh extraction}
Typically, neural SDF representation requires sampling a dense grid of 3D points and evaluate their signed distances before applying Marching Cubes~\cite{MarchingCubes} for mesh extraction. Since in our representation, the shape is defined to be strictly within each GC, we can speed up the inference by skipping points that are outside all GCs. This is achieved by pre-filtering sampling points with bounding boxes of GCs, and computing the relative coordinates in each GC to determine whether a sampling point is inside the GC. More details can be found in the supplementary material.
When a sampling point is inside multiple GCs, we take the minimum value of all the predicted signed distances, which essentially unions all predicted parts in all GCs into a single shape.

\begin{figure}
    \centering
    \includegraphics[width=\linewidth]{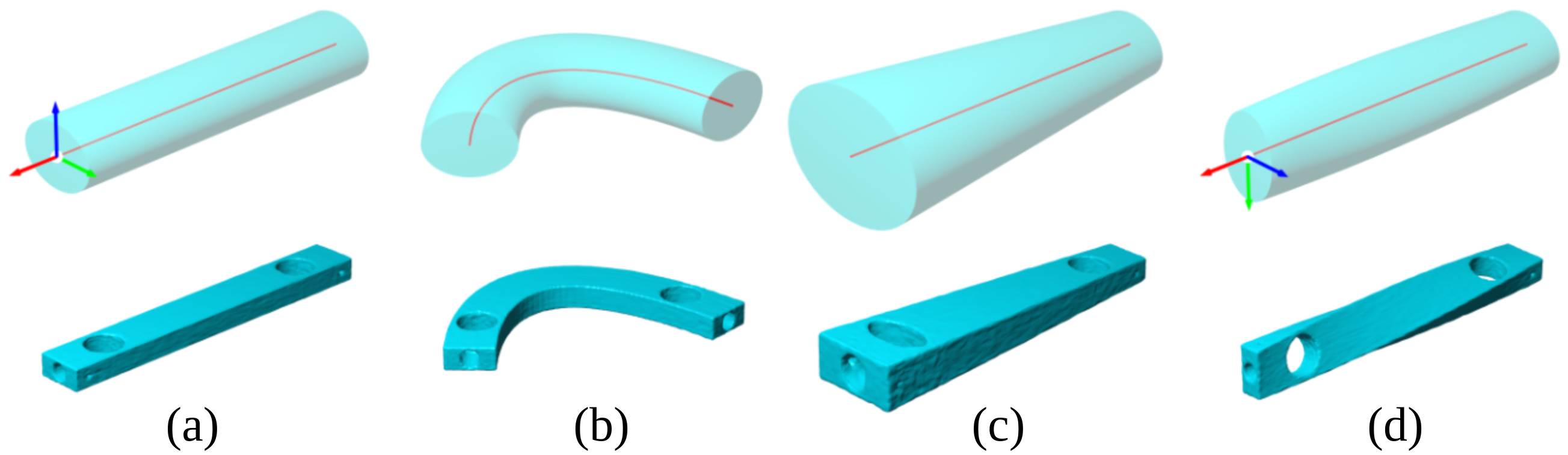}
    \caption{Illustration of shape deformation with GC. (a) The input shape and its covering GC. (b) Changing the shape of the central curve. (c) Increasing the size of one side of the GC. (d) Rotating one side of the GC. }
    \label{fig:3_method_manipulation}
    \vspace{-3mm}
\end{figure}

\parahead{Shape editing}
After fitting, we can easily deform the shape by controlling the shape of the GC, e.g., changing the shape of the curve, and changing the size and the orientation of the key frames, as shown in Fig.~\ref{fig:3_method_manipulation}.

In addition, we can achieve part mixing by blending different parts together. For instance, we can replace a portion of part $A$ with another part $B$ by interpolating their neural features along $A$'s GC curve. Suppose $A$ is represented by global feature vector $z_A$ and $B$ by $z_B$, to obtain the signed distance of point $\Bar{\Point} = (t,a,b)$ in relative coordinates, we can first compute $A$'s feature at its curve point $\Curve_A(t)$, i.e., $\Feat_A(t) = t \MLP_0(z_A) + (1-t) \MLP_1(z_A)$ and similarly for $\Feat_B(t)$. The the signed distance of $\Bar{\Point}$ is given by $h(t,a,b, \textit{blend}(g(\Feat_A(t)),g(\Feat_B(t)),t) )$, where $\textit{blend}(u,v,t)$ is a function to control how $u$ and $v$ are blended with respect to $t$. $\textit{blend}$ can be a variety of functions defined by the user, e.g., simple linear blending $\textit{blend}(u,v,t) = t v + (1-t) u$. Some part mixing results are shown in Fig.~\ref{fig:4_exp_mix_part}.

\section{Experiments}

\subsection{Representation Power}
\begin{figure*}[h]
    \centering
    \includegraphics[width=\linewidth]{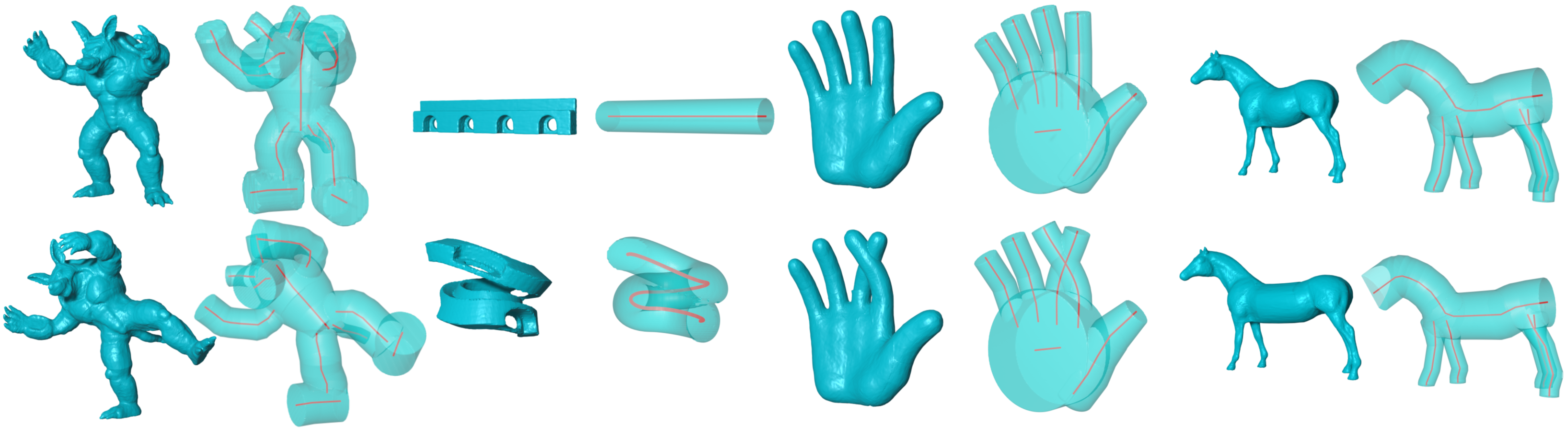}
    \caption{Shape manipulation by changing central curves in the GCs.}
    \label{fig:4_exp_transform_curve}
\end{figure*}

\begin{figure}
    \centering
    \includegraphics[width=1.\linewidth]{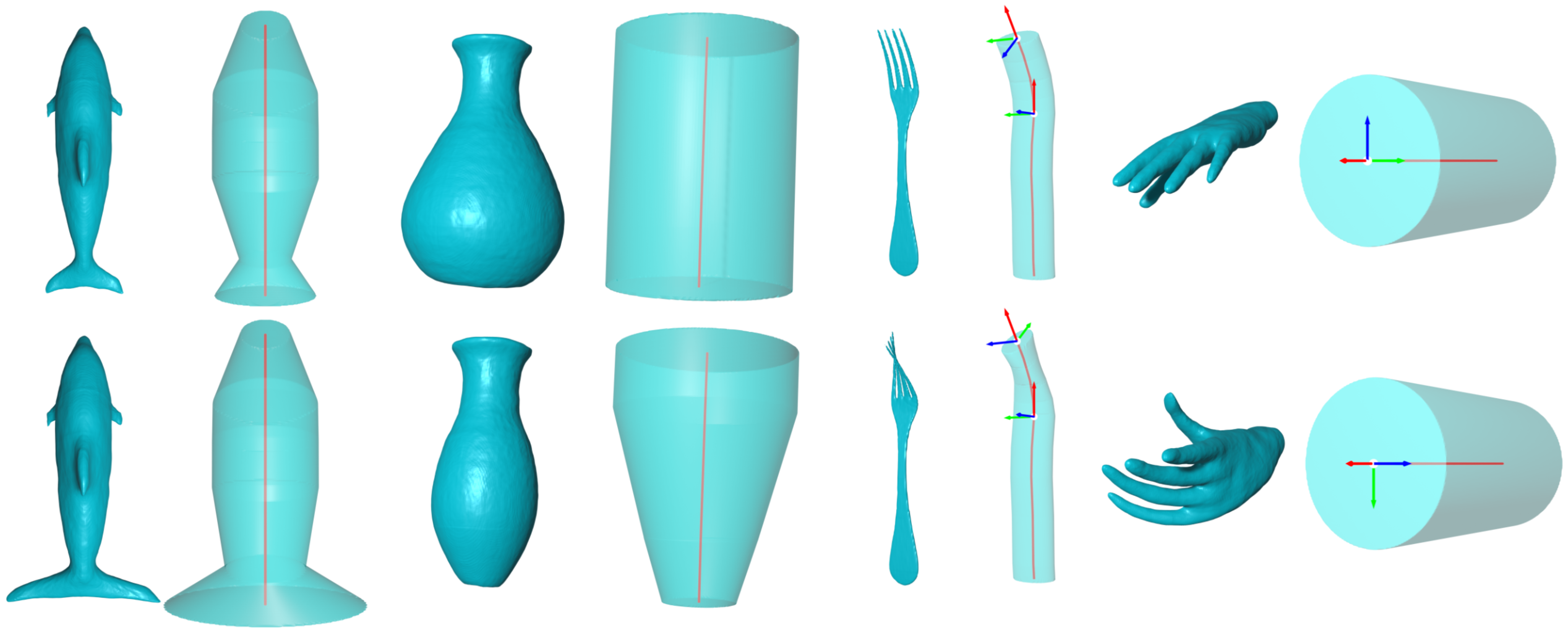}
    \caption{Shape manipulation by changing the oval radii and local frames of the profiles in the GCs.}
    \label{fig:4_exp_scale_tilt}
\end{figure}

\begin{figure*}
    \centering
    \includegraphics[width=0.8\linewidth]{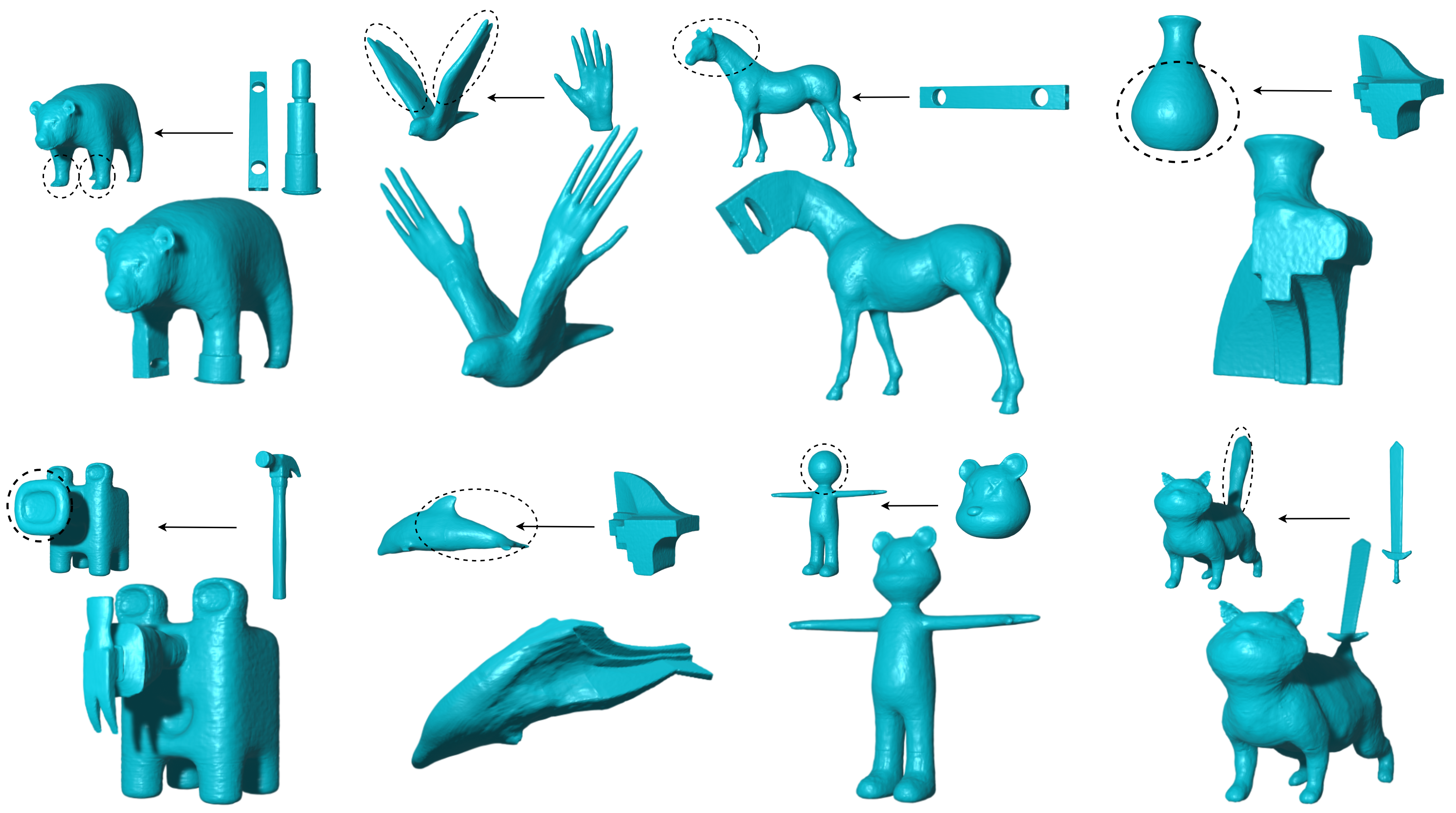}
    \caption{Part mixing by neural feature blending.}
    \label{fig:4_exp_mix_part}
\end{figure*}

To demonstrate the representation power of \ReprName, we train the auto-decoder on a dataset of 50 shapes selected from PSB~\cite{PSB_dataset} and Objaverse~\cite{objaverse}.
The shapes are from various categories, including both organic and man-made objects. We manually annotate GCs for each shape, while ensuring GCs can fully cover the entire object. 
All shapes are normalized to $[-0.8,0.8]^3$. Then we proceed to randomly sample $100,000$ points inside the union of GCs and compute their signed distances for training the neural SDF. Rendered previews of all 50 shapes and other details can be found in the supplementary material.
After training the auto-decoder as described in Sec.~\ref{sec:method_neural}, we apply Marching Cubes in the world frame at $256^3$ resolution to extract meshes from neural SDF. Visual results can be found in Fig.~\ref{fig:4_exp_transform_curve}, \ref{fig:4_exp_scale_tilt}, and \ref{fig:4_exp_mix_part}.

For quantitative evaluation, we report Chamfer Distance (CD), Hausdorff Distance (HD), and Earth Mover's Distance (EMD) between the reconstructed shape and the original shape, averaged over all 50 shapes. Results can be found in Table~\ref{tab:repr_power}.
As a reference, we employ DeepSDF~\cite{deepsdf} as a baseline method and overfit it on those 50 shapes with the same training setting. Our method achieves overall better performance.
In addition, since our annotated GCs follow the nature structure of the shape, e.g., Fig.~\ref{fig:3_method_skeleton_flexible} bottom, it is interesting to test our method when each shape is represented by only a single GC, e.g., Fig.~\ref{fig:3_method_skeleton_flexible} top.
From Table~\ref{tab:repr_power}, it is observed that our method has similar performance regardless of the number of GCs. 

\begin{table}[t]
    \centering
    \caption{Mean fitting error; lower is better.}
    \begin{tabular}{lccc}
    \hline
     & CD & HD & EMD \\
    \hline
    DeepSDF & 1.100e-04 & 2.747e-02 & 2.515e-03\\
    NGC (1 GC) & \textbf{4.122e-05} & \textbf{2.112e-02} & 2.209e-03 \\
    NGC (Ours) & 5.885e-05 & 2.181e-02 & \textbf{1.921e-03} \\
    \hline
    \end{tabular}
    \label{tab:repr_power}
\end{table}

\begin{figure*}[t]
    \centering
    \includegraphics[width=0.9\linewidth]{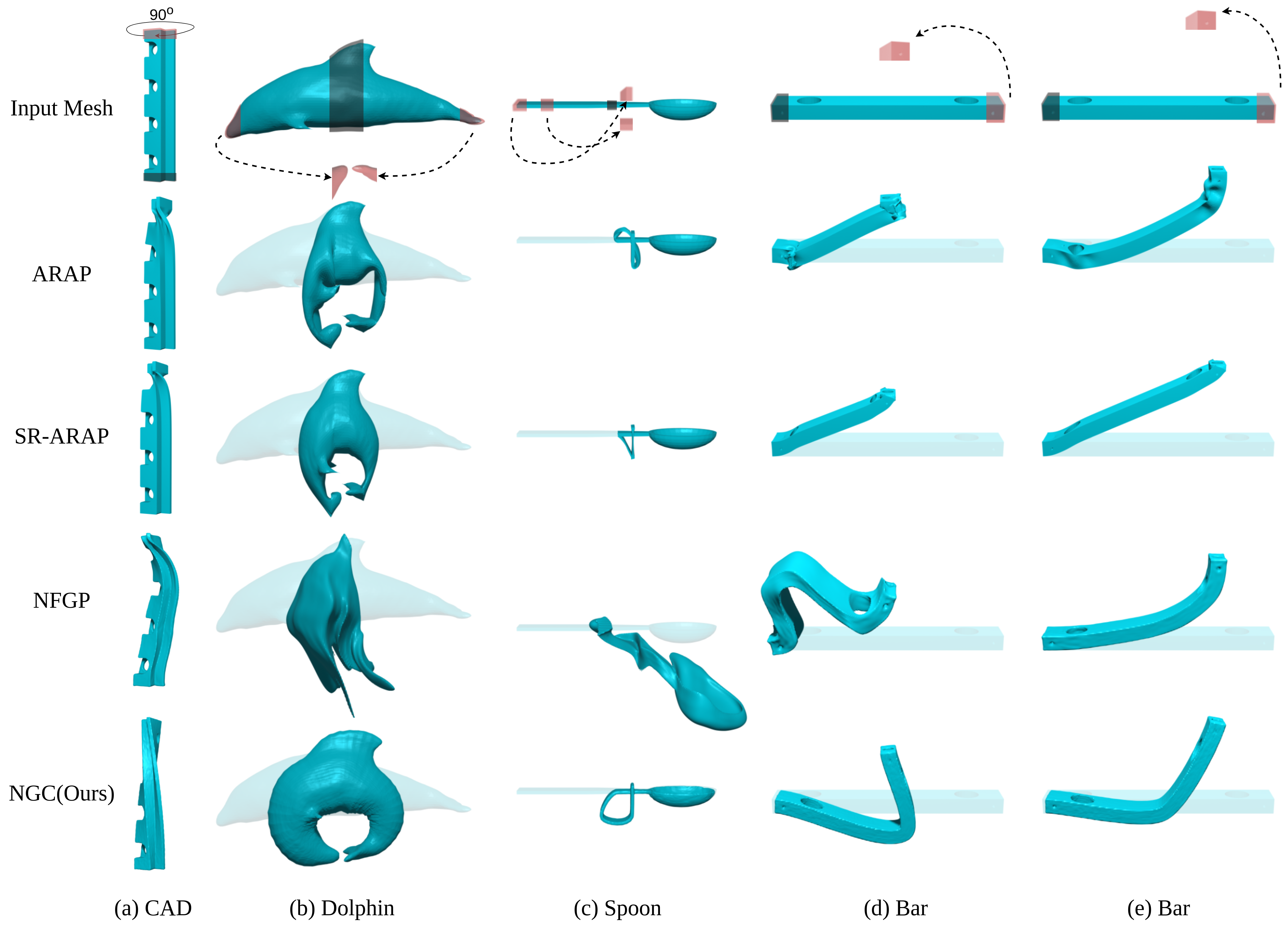}
    \caption{Comparison with ARAP~\cite{ARAP_modeling:2007}, SR-ARAP~\cite{SR_ARAP} and NFGP~\cite{yang2021NFGP}. In the first row, we show five input meshes with deformation constraints. The dark part indicates that this region should be fixed, and the red pairs show the original and target handle points. (d) and (e) are the same input mesh, with only slight difference in the target handle points. All methods can produce reasonable deformations in (e), but only our method is able to handle (d). This and other examples in (a-c) show the robustness and effectiveness of our method. }
    \label{fig:4_exp_deform_comparison}
\end{figure*}

\subsection{Deformation}

%
Given some key handle points on the original shape and their corresponding target positions in the deformed shape, as shown in Fig.~\ref{fig:4_exp_deform_comparison} first row, our \ReprName~ can easily deform the shape by computing a new GC that satisfies the deformation constraints, since in our representation, deforming the GC equates to deforming the shape. It is unnecessary to change the neural features on the GC curve in this case, as we do not wish to change the content of the shape.

To compute the new deformed GC, we first estimate a rigid transformation between the original handle points and the target points, and then convert it to be the transformation between the original and target curve points and local frames on the central curve.
The problem can then be reduced to a Hermitian curve interpolation problem that finds curve interpolating points and tangent vectors (x axis of the local frame).
To be as rigid as possible, we constrain the length of the deformed curve to be the same as the original curve, and keep the radii of the oval profiles to be the same. In general, the curve interpolation with length constraint problem has established solutions in ~\cite{ROULIER199623, curve_length_cons}.

In our experiments, we compare against classic as-rigid-as-possible mesh deformation algorithms ARAP~\cite{ARAP_modeling:2007} and SR-ARAP~\cite{SR_ARAP}, as well as NFGP~\cite{yang2021NFGP} which achieves as-rigid-as-possible deformation on neural SDF via optimization.
%
%
We run these methods with their default settings, as well as our method, on 5 testing cases. Visual comparisons are provided in Fig.~\ref{fig:4_exp_deform_comparison}.
Our method can produce reasonable deformations in challenging cases, while others produce artifacts or even completely fail to preserve the structure of the shape.
%

To quantitatively evaluate the deformation quality, we compute the relative change of volume and area for all methods. Let $V(S)$ be the volume of shape $S$ and $A(S)$ be the area of $S$. Let $\Tilde{S}$ be the deformed shape. The relative change of volume and area can be computed as:
\begin{align*}
    \epsilon_V &= \frac{|V(\Tilde{S}) - V(S)|}{V(S)}, \\
    \epsilon_A &= \frac{|A(\Tilde{S}) - A(S)|}{A(S)}.
\end{align*}
As shown in Table~\ref{tab:vol_err} and ~\ref{tab:area_err}, our method significantly outperforms other methods in terms of volume preservation, and has a decent performance on area preservation.
ARAP~\cite{ARAP_modeling:2007} excels at preserving surface area due to its design of energy functions, as seen in Table~\ref{tab:area_err}. However, its strict area-preservation property often creates deflated or twisted shapes, where the surface should have been shrunk or expanded to accommodate the deformation.
For running speed, our method needs 1.413 seconds on average for each shape, which is comparable to 0.813 second with classic method ARAP~\cite{ARAP_modeling:2007}. In contrast, NFGP~\cite{yang2021NFGP} requires more than 2 hours for optimization.

\begin{table}[t]
    \centering
    \caption{Volume error ($\epsilon_V \times 100$\%); lower is better.}
    \begin{tabular}{lccccc}
    \hline
         & Dolphin & CAD & Spoon & Bar1 & Bar2 \\
    \hline
    ARAP & 45.594 & 8.641 & 26.040 & 20.061 & 5.391 \\
    SR-ARAP & 44.160 & 5.054 & 32.499 & 47.479 & 22.104 \\
    NFGP & 7.607 & 5.576 & 305.573 & 39.004 & 5.517 \\
    NGC (ours) & \textbf{0.325} & \textbf{0.147} & \textbf{1.597} & \textbf{2.614} & \textbf{1.873} \\
    \hline
    \end{tabular}
    \label{tab:vol_err}
\end{table}
\begin{table}[t]
    \centering
    \caption{Area error($\epsilon_A \times 100$\%); lower is better.}
    \begin{tabular}{lccccc}
    \hline
         & Dolphin & CAD & Spoon & Bar1 & Bar2 \\
    \hline
    ARAP & 1.961 & \textbf{1.174} & \textbf{0.468} & \textbf{2.834} & \textbf{2.464} \\
    SR-ARAP & 29.329 & 1.999 & 15.509 & 36.628 & 15.051 \\
    NFGP & 28.047 & 4.318 & 145.673 & 32.607 & 8.408 \\
    NGC (ours) & \textbf{1.415} & 2.696 & 5.829 & 5.471 & 4.730 \\
    \hline
    \end{tabular}
    \label{tab:area_err}
\end{table}

\subsection{Shape Editing}
We provide more shape editing examples enabled by our \ReprName~ representation, including shape deformation by controlling GCs, and part mixing by blending neural features.
In Fig.~\ref{fig:4_exp_transform_curve}, the pose of the armadillo (column 1,2) is changed by rotating curves at the joints of right leg and left arm.
(column 3,4) CAD object is represented by one cylinder and can fit to a curve of arbitrary shape and length. Notice that stretching objects in discretized mesh representation often requires repairing steps like subdivision or remeshing. In contrast, neural SDF does not have such issues.
Another issue is shown in (column 5,6) hand, where crossing fingers can lead to self intersection for discretized meshes, while \ReprName~ simply obtain the boolean union of two fingers.
Finally, (column 7,8) horse shows an example of local stretching on the belly part while keeping the front and back parts identical, by performing a simple reparameterization on the central curve.

Besides manipulation on curves, one can change the oval radii and local frames on GCs for local scaling and twisting of shapes.
In Fig.~\ref{fig:4_exp_scale_tilt}, (column 1,2) dolphin and (column 3,4) vase are scaled at local parts, while (column 5,6) fork and (column 7,8) hand are twisted from one side with the other side fixed. 

So far, we have explored various shape deformation cases, where the content of the shape stays the same.
Nonetheless, the neural features learned by our networks also enable us to change the content of the shapes, by blending neural features from different shapes and parts.
Fig.~\ref{fig:4_exp_mix_part} shows examples of part mixing, where in each example, one or several parts in the original shape are replaced by parts from other shapes.
This is achieved by blending neural features of the original part and the replacement part along the GC curve, as detailed in Sec.~\ref{sec:method_neural}.
The user can adjust the $\textit{blend}$ function to control the range of the mixing area to achieve a smooth blending between two parts.
In our experiments, we use truncated linear function for linear transition of features, while other types of functions can also be used to create different effects. More implementation details of manipulation can be found in the supplementary material.

\begin{figure}
    \centering
    \includegraphics[width=1.\linewidth]{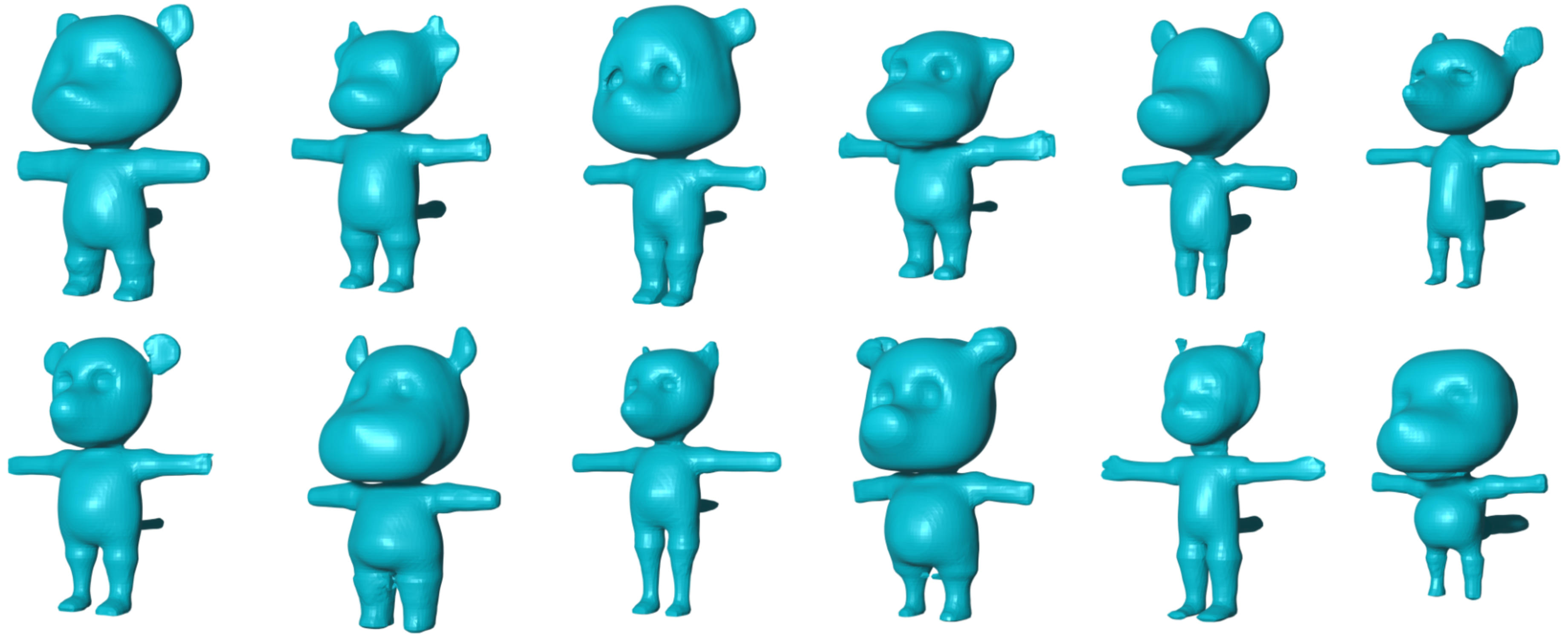}
    \caption{Shape generation when trained on RaBit~\cite{luo2023rabit} dataset.}
    \label{fig:4_exp_generation}
\end{figure}

\subsection{Shape Generation}

We can apply our \ReprName~ representation in a generative model to achieve shape generation.
For each shape, we use one global shape latent code to encode both the curve parameters and \ReprName~ features.
Then we can train an auto-decoder to obtain the latent shape space.
We train our model on RaBit~\cite{luo2023rabit} dataset, as it contains cartoon animal characters in T-pose with skeletons, which is convenient for automatic generation of GCs in our data processing step.
Finally, we train a latent diffusion model on this feature space, with the same setting in EDM~\cite{Karras2022edm}. Some generated shapes are shown in Fig.~\ref{fig:4_exp_generation}, and the results show variations in size of the body and type of the head for random generation. More details can be found in the supplementary material. 

\subsection{Discussion}

\begin{figure}
    \centering
    \includegraphics[width=\linewidth]{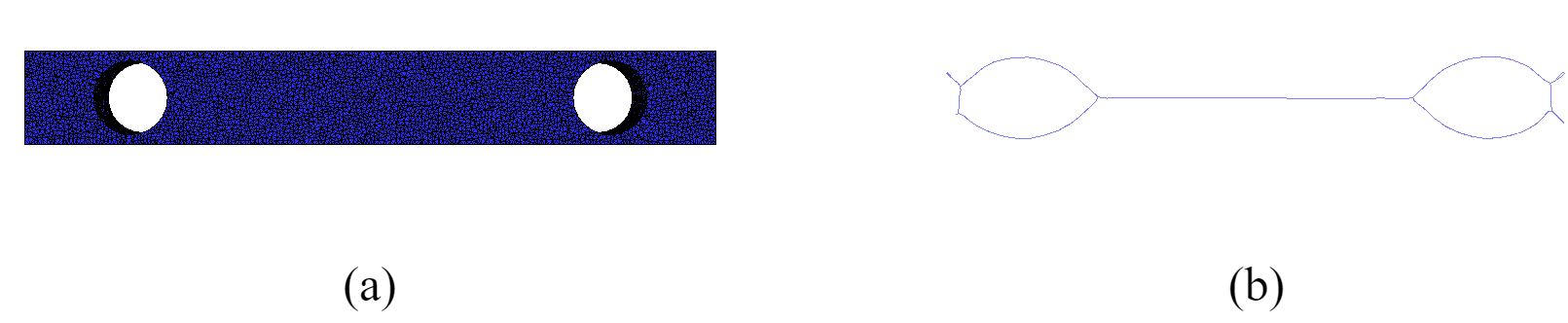}
    \caption{
    Failed skeleton prediction with the traditional skeleton-extraction algorithm. (a) input mesh. (b) predicted skeleton curves from meso skeleton~\cite{meso_skeleton}.
    }
    \label{fig:supp_meso_skeleton}
\end{figure}
\begin{figure}
    \centering
    \includegraphics[width=\linewidth]{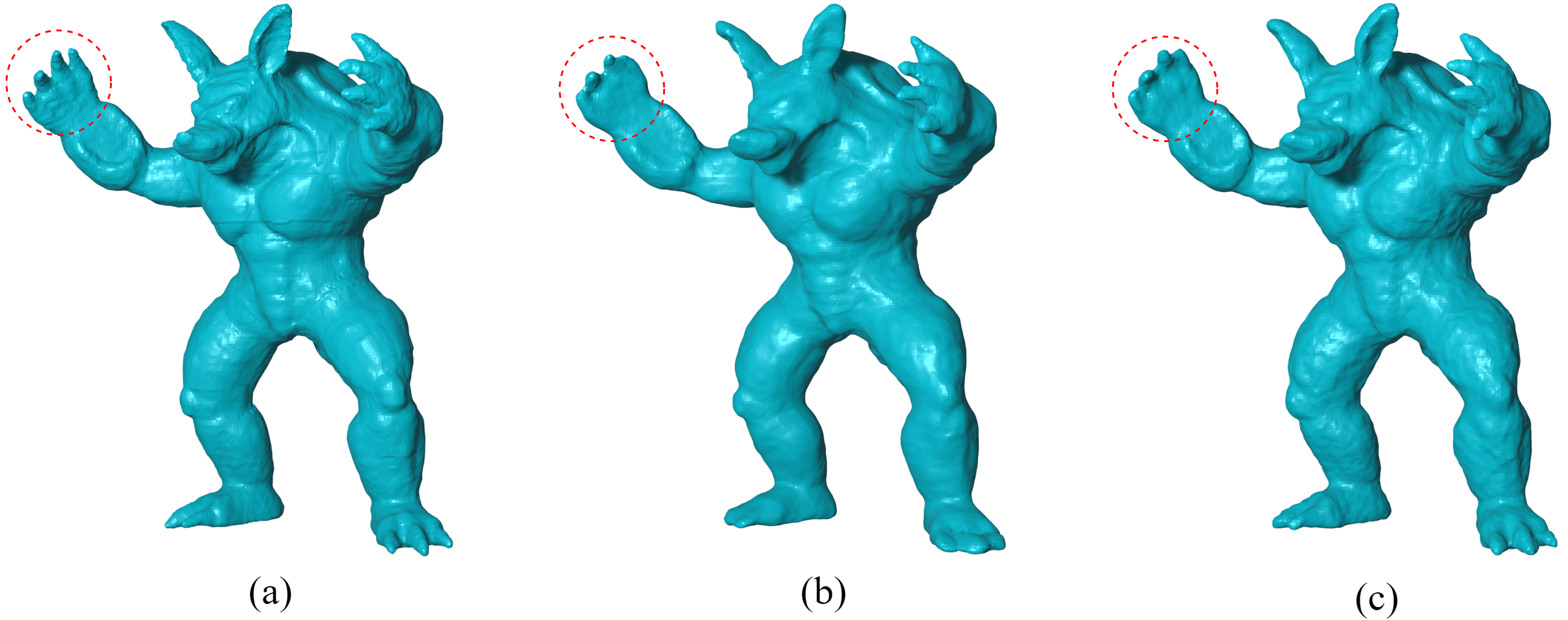}
    \caption{
    Results of armadillo reconstruction for different methods. (a) NGC with multiple GCs. (b) NGC with one GC. (c) DeepSDF.
    }
    \label{fig:supp_repr_power}
\end{figure}

\parahead{Failed automatic skeleton prediction}
For a given triangle mesh, the predicted skeleton curves can be very complicated if we use traditional skeleton-extraction algorithms. In Fig.~\ref{fig:supp_meso_skeleton}, we use meso-skeleton~\cite{meso_skeleton} with CGAL implementation for skeleton extraction, and it produces unsatisfying skeleton curves for users to control. In this case, only one center line is sufficient to represent the shape with \ReprName. Finally, we choose to mannually annotate skeleton curves for all 50 shapes, which can be seen in the supplementary material.

\parahead{Fine details reconstruction}
In Fig.~\ref{fig:supp_repr_power}, we show the case of armadillo reconstruction, where NGC with multiple GCs has better performance on reconstruction of the hand than DeepSDF~\cite{deepsdf} and NGC with single NGC. Intuitively, multiple GCs (see the supplementary material for visualization of the annotated GCs) are provided to decompose the armadillo into several simple parts, which could be better approximated by the neural network.
\section{Conclusion}
In this work, we propose neural generalized cylinder (\ReprName) for controllable shape modeling. By defining the neural SDF on the relative coordinate system of the GC, the shape in \ReprName~can relatively deform with changing of its GC. As a result, the neural SDF can be deformed to expected shapes with explicit constraints. We also compare with classic methods for ARAP deformation, and the results show the superiority of our method qualitatively and quantitatively. Moreover, multiple shapes can be blended together by simply blending their neural features in \ReprName.

\parahead{Limitation and future work}
The main limitation of our method lies in the necessity for manual annotation of curves or skeletons of shapes to facilitate precise deformation or manipulation.
Thus we will seek a suitable skeleton extraction algorithm for \ReprName~ in the future. 
Considering the flexibility of skeleton definition for \ReprName, the skeleton should be as simple as possible for the expected control from users. 
There can also be issues with shape blending where the transitions are not smooth, especially with very different shapes like the dolphin and the fandisk shown in Fig.~\ref{fig:4_exp_mix_part}.
It would be beneficial to explore the incorporation of geometric constraints during the blending process to mitigate these issues. 
Another interesting direction of future work is to further extend the relative coordinate system of \ReprName~ to the full skeleton of a shape.

\begin{acks}
    The work was supported in part by Guangdong Provincial Outstanding Youth Fund(No. 2023B1515020055), the Basic Research Project No. HZQB-KCZYZ-2021067 of Hetao Shenzhen-HK S\&T Cooperation Zone, NSFC-61931024, and Shenzhen Science and Technology Program No. JCYJ20220530143604010. It is also partly supported by  the National Key R\&D Program of China with grant No. 2018YFB1800800, by Shenzhen Outstanding Talents Training Fund 202002, by Guangdong Research Projects No. 2017ZT07X152 and No. 2019CX01X104,by Key Area R\&D Program of Guangdong Province (Grant No. 2018B030338001), by the Guangdong Provincial Key Laboratory of Future Networks of Intelligence (Grant No. 2022B1212010001), by Shenzhen Key Laboratory of Big Data and Artificial Intelligence (Grant No. ZDSYS201707251409055), and by NSFC (62322207).  
\end{acks}
\bibliographystyle{ACM-Reference-Format}
\bibliography{reference}


\begin{thebibliography}{60}


\ifx \showCODEN    \undefined \def \showCODEN     #1{\unskip}     \fi
\ifx \showDOI      \undefined \def \showDOI       #1{#1}\fi
\ifx \showISBNx    \undefined \def \showISBNx     #1{\unskip}     \fi
\ifx \showISBNxiii \undefined \def \showISBNxiii  #1{\unskip}     \fi
\ifx \showISSN     \undefined \def \showISSN      #1{\unskip}     \fi
\ifx \showLCCN     \undefined \def \showLCCN      #1{\unskip}     \fi
\ifx \shownote     \undefined \def \shownote      #1{#1}          \fi
\ifx \showarticletitle \undefined \def \showarticletitle #1{#1}   \fi
\ifx \showURL      \undefined \def \showURL       {\relax}        \fi
\providecommand\bibfield[2]{#2}
\providecommand\bibinfo[2]{#2}
\providecommand\natexlab[1]{#1}
\providecommand\showeprint[2][]{arXiv:#2}

\bibitem[Aigerman et~al\mbox{.}(2022)]%
        {aigerman2022neural}
\bibfield{author}{\bibinfo{person}{Noam Aigerman}, \bibinfo{person}{Kunal Gupta}, \bibinfo{person}{Vladimir~G Kim}, \bibinfo{person}{Siddhartha Chaudhuri}, \bibinfo{person}{Jun Saito}, {and} \bibinfo{person}{Thibault Groueix}.} \bibinfo{year}{2022}\natexlab{}.
\newblock \showarticletitle{Neural jacobian fields: learning intrinsic mappings of arbitrary meshes}.
\newblock \bibinfo{journal}{\emph{ACM Transactions on Graphics (TOG)}} \bibinfo{volume}{41}, \bibinfo{number}{4} (\bibinfo{year}{2022}), \bibinfo{pages}{1--17}.
\newblock


\bibitem[Alexa et~al\mbox{.}(2000)]%
        {alexa2000rigid}
\bibfield{author}{\bibinfo{person}{Marc Alexa}, \bibinfo{person}{Daniel Cohen-Or}, {and} \bibinfo{person}{David Levin}.} \bibinfo{year}{2000}\natexlab{}.
\newblock \showarticletitle{As-rigid-as-possible shape interpolation}. In \bibinfo{booktitle}{\emph{Proceedings of the 27th annual conference on Computer graphics and interactive techniques}}. \bibinfo{pages}{157--164}.
\newblock


\bibitem[Aujay et~al\mbox{.}(2007)]%
        {aujay2007harmonic}
\bibfield{author}{\bibinfo{person}{Gr{\'e}goire Aujay}, \bibinfo{person}{Franck H{\'e}troy}, \bibinfo{person}{Francis Lazarus}, {and} \bibinfo{person}{Christine Depraz}.} \bibinfo{year}{2007}\natexlab{}.
\newblock \showarticletitle{Harmonic skeleton for realistic character animation}. In \bibinfo{booktitle}{\emph{SCA 2007-ACM-SIGGRAPH/Eurographics Symposium on Computer Animation}}. Eurographics Association, \bibinfo{pages}{151--160}.
\newblock


\bibitem[B\ae{}rentzen et~al\mbox{.}(2014)]%
        {PAM2014}
\bibfield{author}{\bibinfo{person}{J.~Andreas B\ae{}rentzen}, \bibinfo{person}{Rinat Abdrashitov}, {and} \bibinfo{person}{Karan Singh}.} \bibinfo{year}{2014}\natexlab{}.
\newblock \showarticletitle{Interactive shape modeling using a skeleton-mesh co-representation}.
\newblock \bibinfo{journal}{\emph{ACM Trans. Graph.}} \bibinfo{volume}{33}, \bibinfo{number}{4}, Article \bibinfo{articleno}{132} (\bibinfo{date}{jul} \bibinfo{year}{2014}), \bibinfo{numpages}{10}~pages.
\newblock
\showISSN{0730-0301}
\urldef\tempurl%
\url{https://doi.org/10.1145/2601097.2601226}
\showDOI{\tempurl}


\bibitem[Baran and Popovi{\'c}(2007)]%
        {baran2007automatic}
\bibfield{author}{\bibinfo{person}{Ilya Baran} {and} \bibinfo{person}{Jovan Popovi{\'c}}.} \bibinfo{year}{2007}\natexlab{}.
\newblock \showarticletitle{Automatic rigging and animation of 3d characters}.
\newblock \bibinfo{journal}{\emph{ACM Transactions on graphics (TOG)}} \bibinfo{volume}{26}, \bibinfo{number}{3} (\bibinfo{year}{2007}), \bibinfo{pages}{72--es}.
\newblock


\bibitem[Botsch and Sorkine(2007)]%
        {botsch2007linear}
\bibfield{author}{\bibinfo{person}{Mario Botsch} {and} \bibinfo{person}{Olga Sorkine}.} \bibinfo{year}{2007}\natexlab{}.
\newblock \showarticletitle{On linear variational surface deformation methods}.
\newblock \bibinfo{journal}{\emph{IEEE transactions on visualization and computer graphics}} \bibinfo{volume}{14}, \bibinfo{number}{1} (\bibinfo{year}{2007}), \bibinfo{pages}{213--230}.
\newblock


\bibitem[Chen et~al\mbox{.}(2013)]%
        {Chen2013_3sweep}
\bibfield{author}{\bibinfo{person}{Tao Chen}, \bibinfo{person}{Zhe Zhu}, \bibinfo{person}{Ariel Shamir}, \bibinfo{person}{Shi-Min Hu}, {and} \bibinfo{person}{Daniel Cohen-Or}.} \bibinfo{year}{2013}\natexlab{}.
\newblock \showarticletitle{3-Sweep: extracting editable objects from a single photo}.
\newblock \bibinfo{journal}{\emph{ACM Trans. Graph.}} \bibinfo{volume}{32}, \bibinfo{number}{6}, Article \bibinfo{articleno}{195} (\bibinfo{date}{nov} \bibinfo{year}{2013}), \bibinfo{numpages}{10}~pages.
\newblock
\showISSN{0730-0301}
\urldef\tempurl%
\url{https://doi.org/10.1145/2508363.2508378}
\showDOI{\tempurl}


\bibitem[Chen et~al\mbox{.}(2009)]%
        {PSB_dataset}
\bibfield{author}{\bibinfo{person}{Xiaobai Chen}, \bibinfo{person}{Aleksey Golovinskiy}, {and} \bibinfo{person}{Thomas Funkhouser}.} \bibinfo{year}{2009}\natexlab{}.
\newblock \showarticletitle{A Benchmark for {3D} Mesh Segmentation}.
\newblock \bibinfo{journal}{\emph{ACM Transactions on Graphics (Proc. SIGGRAPH)}} \bibinfo{volume}{28}, \bibinfo{number}{3} (\bibinfo{date}{Aug.} \bibinfo{year}{2009}).
\newblock


\bibitem[Chen et~al\mbox{.}(2021)]%
        {chen2021snarf}
\bibfield{author}{\bibinfo{person}{Xu Chen}, \bibinfo{person}{Yufeng Zheng}, \bibinfo{person}{Michael~J Black}, \bibinfo{person}{Otmar Hilliges}, {and} \bibinfo{person}{Andreas Geiger}.} \bibinfo{year}{2021}\natexlab{}.
\newblock \showarticletitle{Snarf: Differentiable forward skinning for animating non-rigid neural implicit shapes}. In \bibinfo{booktitle}{\emph{Proceedings of the IEEE/CVF International Conference on Computer Vision}}. \bibinfo{pages}{11594--11604}.
\newblock


\bibitem[Chen and Zhang(2018)]%
        {IMnet}
\bibfield{author}{\bibinfo{person}{Zhiqin Chen} {and} \bibinfo{person}{Hao Zhang}.} \bibinfo{year}{2018}\natexlab{}.
\newblock \showarticletitle{Learning Implicit Fields for Generative Shape Modeling}.
\newblock \bibinfo{journal}{\emph{CoRR}}  \bibinfo{volume}{abs/1812.02822} (\bibinfo{year}{2018}).
\newblock
\showeprint[arXiv]{1812.02822}
\urldef\tempurl%
\url{http://arxiv.org/abs/1812.02822}
\showURL{%
\tempurl}


\bibitem[Deitke et~al\mbox{.}(2023)]%
        {objaverse}
\bibfield{author}{\bibinfo{person}{Matt Deitke}, \bibinfo{person}{Dustin Schwenk}, \bibinfo{person}{Jordi Salvador}, \bibinfo{person}{Luca Weihs}, \bibinfo{person}{Oscar Michel}, \bibinfo{person}{Eli VanderBilt}, \bibinfo{person}{Ludwig Schmidt}, \bibinfo{person}{Kiana Ehsani}, \bibinfo{person}{Aniruddha Kembhavi}, {and} \bibinfo{person}{Ali Farhadi}.} \bibinfo{year}{2023}\natexlab{}.
\newblock \showarticletitle{Objaverse: A Universe of Annotated 3D Objects}. In \bibinfo{booktitle}{\emph{Proceedings of the IEEE/CVF Conference on Computer Vision and Pattern Recognition}}. \bibinfo{pages}{13142--13153}.
\newblock


\bibitem[Funkhouser et~al\mbox{.}(2004)]%
        {funkhouser2004modeling}
\bibfield{author}{\bibinfo{person}{Thomas Funkhouser}, \bibinfo{person}{Michael Kazhdan}, \bibinfo{person}{Philip Shilane}, \bibinfo{person}{Patrick Min}, \bibinfo{person}{William Kiefer}, \bibinfo{person}{Ayellet Tal}, \bibinfo{person}{Szymon Rusinkiewicz}, {and} \bibinfo{person}{David Dobkin}.} \bibinfo{year}{2004}\natexlab{}.
\newblock \showarticletitle{Modeling by example}.
\newblock \bibinfo{journal}{\emph{ACM transactions on graphics (TOG)}} \bibinfo{volume}{23}, \bibinfo{number}{3} (\bibinfo{year}{2004}), \bibinfo{pages}{652--663}.
\newblock


\bibitem[Groueix et~al\mbox{.}(2018)]%
        {groueix2018papier}
\bibfield{author}{\bibinfo{person}{Thibault Groueix}, \bibinfo{person}{Matthew Fisher}, \bibinfo{person}{Vladimir~G Kim}, \bibinfo{person}{Bryan~C Russell}, {and} \bibinfo{person}{Mathieu Aubry}.} \bibinfo{year}{2018}\natexlab{}.
\newblock \showarticletitle{A papier-m{\^a}ch{\'e} approach to learning 3d surface generation}. In \bibinfo{booktitle}{\emph{Proceedings of the IEEE conference on computer vision and pattern recognition}}. \bibinfo{pages}{216--224}.
\newblock


\bibitem[Haque et~al\mbox{.}(2023)]%
        {haque2023instruct}
\bibfield{author}{\bibinfo{person}{Ayaan Haque}, \bibinfo{person}{Matthew Tancik}, \bibinfo{person}{Alexei~A Efros}, \bibinfo{person}{Aleksander Holynski}, {and} \bibinfo{person}{Angjoo Kanazawa}.} \bibinfo{year}{2023}\natexlab{}.
\newblock \showarticletitle{Instruct-nerf2nerf: Editing 3d scenes with instructions}. In \bibinfo{booktitle}{\emph{Proceedings of the IEEE/CVF International Conference on Computer Vision}}. \bibinfo{pages}{19740--19750}.
\newblock


\bibitem[Hertz et~al\mbox{.}(2022)]%
        {hertz2022spaghetti}
\bibfield{author}{\bibinfo{person}{Amir Hertz}, \bibinfo{person}{Or Perel}, \bibinfo{person}{Raja Giryes}, \bibinfo{person}{Olga Sorkine-Hornung}, {and} \bibinfo{person}{Daniel Cohen-Or}.} \bibinfo{year}{2022}\natexlab{}.
\newblock \showarticletitle{Spaghetti: Editing implicit shapes through part aware generation}.
\newblock \bibinfo{journal}{\emph{ACM Transactions on Graphics (TOG)}} \bibinfo{volume}{41}, \bibinfo{number}{4} (\bibinfo{year}{2022}), \bibinfo{pages}{1--20}.
\newblock


\bibitem[Jacobson et~al\mbox{.}(2011)]%
        {jacobson2011bounded}
\bibfield{author}{\bibinfo{person}{Alec Jacobson}, \bibinfo{person}{Ilya Baran}, \bibinfo{person}{Jovan Popovic}, {and} \bibinfo{person}{Olga Sorkine}.} \bibinfo{year}{2011}\natexlab{}.
\newblock \showarticletitle{Bounded biharmonic weights for real-time deformation.}
\newblock \bibinfo{journal}{\emph{ACM Trans. Graph.}} \bibinfo{volume}{30}, \bibinfo{number}{4} (\bibinfo{year}{2011}), \bibinfo{pages}{78}.
\newblock


\bibitem[Jakab et~al\mbox{.}(2021)]%
        {jakab2021keypointdeformer}
\bibfield{author}{\bibinfo{person}{Tomas Jakab}, \bibinfo{person}{Richard Tucker}, \bibinfo{person}{Ameesh Makadia}, \bibinfo{person}{Jiajun Wu}, \bibinfo{person}{Noah Snavely}, {and} \bibinfo{person}{Angjoo Kanazawa}.} \bibinfo{year}{2021}\natexlab{}.
\newblock \showarticletitle{Keypointdeformer: Unsupervised 3d keypoint discovery for shape control}. In \bibinfo{booktitle}{\emph{Proceedings of the IEEE/CVF Conference on Computer Vision and Pattern Recognition}}. \bibinfo{pages}{12783--12792}.
\newblock


\bibitem[Joshi et~al\mbox{.}(2007)]%
        {joshi2007harmonic}
\bibfield{author}{\bibinfo{person}{Pushkar Joshi}, \bibinfo{person}{Mark Meyer}, \bibinfo{person}{Tony DeRose}, \bibinfo{person}{Brian Green}, {and} \bibinfo{person}{Tom Sanocki}.} \bibinfo{year}{2007}\natexlab{}.
\newblock \showarticletitle{Harmonic coordinates for character articulation}.
\newblock \bibinfo{journal}{\emph{ACM transactions on graphics (TOG)}} \bibinfo{volume}{26}, \bibinfo{number}{3} (\bibinfo{year}{2007}), \bibinfo{pages}{71--es}.
\newblock


\bibitem[Ju et~al\mbox{.}(2023)]%
        {ju2023mean}
\bibfield{author}{\bibinfo{person}{Tao Ju}, \bibinfo{person}{Scott Schaefer}, {and} \bibinfo{person}{Joe Warren}.} \bibinfo{year}{2023}\natexlab{}.
\newblock \showarticletitle{Mean value coordinates for closed triangular meshes}.
\newblock In \bibinfo{booktitle}{\emph{Seminal Graphics Papers: Pushing the Boundaries, Volume 2}}. \bibinfo{pages}{223--228}.
\newblock


\bibitem[Karras et~al\mbox{.}(2022)]%
        {Karras2022edm}
\bibfield{author}{\bibinfo{person}{Tero Karras}, \bibinfo{person}{Miika Aittala}, \bibinfo{person}{Timo Aila}, {and} \bibinfo{person}{Samuli Laine}.} \bibinfo{year}{2022}\natexlab{}.
\newblock \showarticletitle{Elucidating the Design Space of Diffusion-Based Generative Models}. In \bibinfo{booktitle}{\emph{Proc. NeurIPS}}.
\newblock


\bibitem[Kho and Garland(2005)]%
        {kho2005sketching}
\bibfield{author}{\bibinfo{person}{Youngihn Kho} {and} \bibinfo{person}{Michael Garland}.} \bibinfo{year}{2005}\natexlab{}.
\newblock \showarticletitle{Sketching mesh deformations}. In \bibinfo{booktitle}{\emph{Proceedings of the 2005 symposium on Interactive 3D graphics and games}}. \bibinfo{pages}{147--154}.
\newblock


\bibitem[Levi and Gotsman(2015)]%
        {SR_ARAP}
\bibfield{author}{\bibinfo{person}{Zohar Levi} {and} \bibinfo{person}{Craig Gotsman}.} \bibinfo{year}{2015}\natexlab{}.
\newblock \showarticletitle{Smooth Rotation Enhanced As-Rigid-As-Possible Mesh Animation}.
\newblock \bibinfo{journal}{\emph{IEEE Transactions on Visualization and Computer Graphics}} \bibinfo{volume}{21}, \bibinfo{number}{2} (\bibinfo{year}{2015}), \bibinfo{pages}{264--277}.
\newblock
\urldef\tempurl%
\url{https://doi.org/10.1109/TVCG.2014.2359463}
\showDOI{\tempurl}


\bibitem[Levi and Levin(2014)]%
        {deformation_IRBF}
\bibfield{author}{\bibinfo{person}{Z. Levi} {and} \bibinfo{person}{D. Levin}.} \bibinfo{year}{2014}\natexlab{}.
\newblock \showarticletitle{Shape Deformation via Interior RBF}.
\newblock \bibinfo{journal}{\emph{IEEE Transactions on Visualization \& amp; Computer Graphics}} \bibinfo{volume}{20}, \bibinfo{number}{07} (\bibinfo{date}{jul} \bibinfo{year}{2014}), \bibinfo{pages}{1062--1075}.
\newblock
\showISSN{1941-0506}
\urldef\tempurl%
\url{https://doi.org/10.1109/TVCG.2013.255}
\showDOI{\tempurl}


\bibitem[Li et~al\mbox{.}(2010)]%
        {arterial_snakes2010}
\bibfield{author}{\bibinfo{person}{Guo Li}, \bibinfo{person}{Ligang Liu}, \bibinfo{person}{Hanlin Zheng}, {and} \bibinfo{person}{Niloy~J. Mitra}.} \bibinfo{year}{2010}\natexlab{}.
\newblock \showarticletitle{Analysis, reconstruction and manipulation using arterial snakes}.
\newblock \bibinfo{journal}{\emph{ACM Trans. Graph.}} \bibinfo{volume}{29}, \bibinfo{number}{6}, Article \bibinfo{articleno}{152} (\bibinfo{date}{dec} \bibinfo{year}{2010}), \bibinfo{numpages}{10}~pages.
\newblock
\showISSN{0730-0301}
\urldef\tempurl%
\url{https://doi.org/10.1145/1882261.1866178}
\showDOI{\tempurl}


\bibitem[Li et~al\mbox{.}(2021)]%
        {li2021sp}
\bibfield{author}{\bibinfo{person}{Ruihui Li}, \bibinfo{person}{Xianzhi Li}, \bibinfo{person}{Ka-Hei Hui}, {and} \bibinfo{person}{Chi-Wing Fu}.} \bibinfo{year}{2021}\natexlab{}.
\newblock \showarticletitle{SP-GAN: Sphere-guided 3D shape generation and manipulation}.
\newblock \bibinfo{journal}{\emph{ACM Transactions on Graphics (TOG)}} \bibinfo{volume}{40}, \bibinfo{number}{4} (\bibinfo{year}{2021}), \bibinfo{pages}{1--12}.
\newblock


\bibitem[Lipman et~al\mbox{.}(2008)]%
        {lipman2008green}
\bibfield{author}{\bibinfo{person}{Yaron Lipman}, \bibinfo{person}{David Levin}, {and} \bibinfo{person}{Daniel Cohen-Or}.} \bibinfo{year}{2008}\natexlab{}.
\newblock \showarticletitle{Green coordinates}.
\newblock \bibinfo{journal}{\emph{ACM transactions on graphics (TOG)}} \bibinfo{volume}{27}, \bibinfo{number}{3} (\bibinfo{year}{2008}), \bibinfo{pages}{1--10}.
\newblock


\bibitem[Lipman et~al\mbox{.}(2004)]%
        {lipman2004differential}
\bibfield{author}{\bibinfo{person}{Yaron Lipman}, \bibinfo{person}{Olga Sorkine}, \bibinfo{person}{Daniel Cohen-Or}, \bibinfo{person}{David Levin}, \bibinfo{person}{Christian Rossi}, {and} \bibinfo{person}{Hans-Peter Seidel}.} \bibinfo{year}{2004}\natexlab{}.
\newblock \showarticletitle{Differential coordinates for interactive mesh editing}. In \bibinfo{booktitle}{\emph{Proceedings Shape Modeling Applications, 2004.}} IEEE, \bibinfo{pages}{181--190}.
\newblock


\bibitem[Lorensen and Cline(1987)]%
        {MarchingCubes}
\bibfield{author}{\bibinfo{person}{William~E. Lorensen} {and} \bibinfo{person}{Harvey~E. Cline}.} \bibinfo{year}{1987}\natexlab{}.
\newblock \showarticletitle{Marching cubes: A high resolution 3D surface construction algorithm.}. In \bibinfo{booktitle}{\emph{SIGGRAPH}}, \bibfield{editor}{\bibinfo{person}{Maureen~C. Stone}} (Ed.). \bibinfo{publisher}{ACM}, \bibinfo{pages}{163--169}.
\newblock
\showISBNx{0-89791-227-6}
\urldef\tempurl%
\url{http://dblp.uni-trier.de/db/conf/siggraph/siggraph1987.html#LorensenC87}
\showURL{%
\tempurl}


\bibitem[Luo et~al\mbox{.}(2023)]%
        {luo2023rabit}
\bibfield{author}{\bibinfo{person}{Zhongjin Luo}, \bibinfo{person}{Shengcai Cai}, \bibinfo{person}{Jinguo Dong}, \bibinfo{person}{Ruibo Ming}, \bibinfo{person}{Liangdong Qiu}, \bibinfo{person}{Xiaohang Zhan}, {and} \bibinfo{person}{Xiaoguang Han}.} \bibinfo{year}{2023}\natexlab{}.
\newblock \showarticletitle{RaBit: Parametric Modeling of 3D Biped Cartoon Characters with a Topological-consistent Dataset}. In \bibinfo{booktitle}{\emph{Proceedings of the IEEE/CVF Conference on Computer Vision and Pattern Recognition (CVPR)}}.
\newblock


\bibitem[Ma et~al\mbox{.}(2018)]%
        {ma2018skeleton}
\bibfield{author}{\bibinfo{person}{Ruibin Ma}, \bibinfo{person}{Qingyu Zhao}, \bibinfo{person}{Rui Wang}, \bibinfo{person}{James~N Damon}, \bibinfo{person}{Julian~G Rosenman}, {and} \bibinfo{person}{Stephen~M Pizer}.} \bibinfo{year}{2018}\natexlab{}.
\newblock \showarticletitle{Skeleton-based Generalized Cylinder Deformation under the Relative Curvature Condition.}. In \bibinfo{booktitle}{\emph{PG (Short Papers and Posters)}}. \bibinfo{pages}{37--40}.
\newblock


\bibitem[Mescheder et~al\mbox{.}(2019)]%
        {occnet}
\bibfield{author}{\bibinfo{person}{Lars Mescheder}, \bibinfo{person}{Michael Oechsle}, \bibinfo{person}{Michael Niemeyer}, \bibinfo{person}{Sebastian Nowozin}, {and} \bibinfo{person}{Andreas Geiger}.} \bibinfo{year}{2019}\natexlab{}.
\newblock \showarticletitle{Occupancy networks: Learning 3d reconstruction in function space}. In \bibinfo{booktitle}{\emph{Proceedings of the IEEE/CVF conference on computer vision and pattern recognition}}. \bibinfo{pages}{4460--4470}.
\newblock


\bibitem[Museth et~al\mbox{.}(2002)]%
        {museth2002level}
\bibfield{author}{\bibinfo{person}{Ken Museth}, \bibinfo{person}{David~E Breen}, \bibinfo{person}{Ross~T Whitaker}, {and} \bibinfo{person}{Alan~H Barr}.} \bibinfo{year}{2002}\natexlab{}.
\newblock \showarticletitle{Level set surface editing operators}. In \bibinfo{booktitle}{\emph{Proceedings of the 29th annual conference on Computer graphics and interactive techniques}}. \bibinfo{pages}{330--338}.
\newblock


\bibitem[Osher et~al\mbox{.}(2004)]%
        {osher2004level}
\bibfield{author}{\bibinfo{person}{Stanley Osher}, \bibinfo{person}{Ronald Fedkiw}, {and} \bibinfo{person}{K Piechor}.} \bibinfo{year}{2004}\natexlab{}.
\newblock \showarticletitle{Level set methods and dynamic implicit surfaces}.
\newblock \bibinfo{journal}{\emph{Appl. Mech. Rev.}} \bibinfo{volume}{57}, \bibinfo{number}{3} (\bibinfo{year}{2004}), \bibinfo{pages}{B15--B15}.
\newblock


\bibitem[Park et~al\mbox{.}(2019)]%
        {deepsdf}
\bibfield{author}{\bibinfo{person}{Jeong~Joon Park}, \bibinfo{person}{Peter Florence}, \bibinfo{person}{Julian Straub}, \bibinfo{person}{Richard Newcombe}, {and} \bibinfo{person}{Steven Lovegrove}.} \bibinfo{year}{2019}\natexlab{}.
\newblock \showarticletitle{DeepSDF: Learning Continuous Signed Distance Functions for Shape Representation}. In \bibinfo{booktitle}{\emph{The IEEE Conference on Computer Vision and Pattern Recognition (CVPR)}}.
\newblock


\bibitem[Park et~al\mbox{.}(2021)]%
        {park2021nerfies}
\bibfield{author}{\bibinfo{person}{Keunhong Park}, \bibinfo{person}{Utkarsh Sinha}, \bibinfo{person}{Jonathan~T Barron}, \bibinfo{person}{Sofien Bouaziz}, \bibinfo{person}{Dan~B Goldman}, \bibinfo{person}{Steven~M Seitz}, {and} \bibinfo{person}{Ricardo Martin-Brualla}.} \bibinfo{year}{2021}\natexlab{}.
\newblock \showarticletitle{Nerfies: Deformable neural radiance fields}. In \bibinfo{booktitle}{\emph{Proceedings of the IEEE/CVF International Conference on Computer Vision}}. \bibinfo{pages}{5865--5874}.
\newblock


\bibitem[Peng et~al\mbox{.}(2022)]%
        {peng2022cagenerf}
\bibfield{author}{\bibinfo{person}{Yicong Peng}, \bibinfo{person}{Yichao Yan}, \bibinfo{person}{Shengqi Liu}, \bibinfo{person}{Yuhao Cheng}, \bibinfo{person}{Shanyan Guan}, \bibinfo{person}{Bowen Pan}, \bibinfo{person}{Guangtao Zhai}, {and} \bibinfo{person}{Xiaokang Yang}.} \bibinfo{year}{2022}\natexlab{}.
\newblock \showarticletitle{Cagenerf: Cage-based neural radiance field for generalized 3d deformation and animation}.
\newblock \bibinfo{journal}{\emph{Advances in Neural Information Processing Systems}}  \bibinfo{volume}{35} (\bibinfo{year}{2022}), \bibinfo{pages}{31402--31415}.
\newblock


\bibitem[Pumarola et~al\mbox{.}(2021)]%
        {pumarola2021d}
\bibfield{author}{\bibinfo{person}{Albert Pumarola}, \bibinfo{person}{Enric Corona}, \bibinfo{person}{Gerard Pons-Moll}, {and} \bibinfo{person}{Francesc Moreno-Noguer}.} \bibinfo{year}{2021}\natexlab{}.
\newblock \showarticletitle{D-nerf: Neural radiance fields for dynamic scenes}. In \bibinfo{booktitle}{\emph{Proceedings of the IEEE/CVF Conference on Computer Vision and Pattern Recognition}}. \bibinfo{pages}{10318--10327}.
\newblock


\bibitem[Roulier and Piper(1996)]%
        {ROULIER199623}
\bibfield{author}{\bibinfo{person}{John~A. Roulier} {and} \bibinfo{person}{Bruce Piper}.} \bibinfo{year}{1996}\natexlab{}.
\newblock \showarticletitle{Prescribing the length of rational Bézier curves}.
\newblock \bibinfo{journal}{\emph{Computer Aided Geometric Design}} \bibinfo{volume}{13}, \bibinfo{number}{1} (\bibinfo{year}{1996}), \bibinfo{pages}{23--43}.
\newblock
\showISSN{0167-8396}
\urldef\tempurl%
\url{https://doi.org/10.1016/0167-8396(95)00005-4}
\showDOI{\tempurl}


\bibitem[Sederberg and Parry(1986)]%
        {sederberg1986free}
\bibfield{author}{\bibinfo{person}{Thomas~W Sederberg} {and} \bibinfo{person}{Scott~R Parry}.} \bibinfo{year}{1986}\natexlab{}.
\newblock \showarticletitle{Free-form deformation of solid geometric models}. In \bibinfo{booktitle}{\emph{Proceedings of the 13th annual conference on Computer graphics and interactive techniques}}. \bibinfo{pages}{151--160}.
\newblock


\bibitem[Sorkine and Alexa(2007a)]%
        {ARAP_modeling:2007}
\bibfield{author}{\bibinfo{person}{Olga Sorkine} {and} \bibinfo{person}{Marc Alexa}.} \bibinfo{year}{2007}\natexlab{a}.
\newblock \showarticletitle{As-Rigid-As-Possible Surface Modeling}. In \bibinfo{booktitle}{\emph{Proceedings of EUROGRAPHICS/ACM SIGGRAPH Symposium on Geometry Processing}}. \bibinfo{pages}{109--116}.
\newblock


\bibitem[Sorkine and Alexa(2007b)]%
        {sorkine2007rigid}
\bibfield{author}{\bibinfo{person}{Olga Sorkine} {and} \bibinfo{person}{Marc Alexa}.} \bibinfo{year}{2007}\natexlab{b}.
\newblock \showarticletitle{As-rigid-as-possible surface modeling}. In \bibinfo{booktitle}{\emph{Symposium on Geometry processing}}, Vol.~\bibinfo{volume}{4}. Citeseer, \bibinfo{pages}{109--116}.
\newblock


\bibitem[Sorkine et~al\mbox{.}(2004)]%
        {sorkine2004laplacian}
\bibfield{author}{\bibinfo{person}{Olga Sorkine}, \bibinfo{person}{Daniel Cohen-Or}, \bibinfo{person}{Yaron Lipman}, \bibinfo{person}{Marc Alexa}, \bibinfo{person}{Christian R{\"o}ssl}, {and} \bibinfo{person}{H-P Seidel}.} \bibinfo{year}{2004}\natexlab{}.
\newblock \showarticletitle{Laplacian surface editing}. In \bibinfo{booktitle}{\emph{Proceedings of the 2004 Eurographics/ACM SIGGRAPH symposium on Geometry processing}}. \bibinfo{pages}{175--184}.
\newblock


\bibitem[Sumner et~al\mbox{.}(2007)]%
        {sumner2007embedded}
\bibfield{author}{\bibinfo{person}{Robert~W Sumner}, \bibinfo{person}{Johannes Schmid}, {and} \bibinfo{person}{Mark Pauly}.} \bibinfo{year}{2007}\natexlab{}.
\newblock \showarticletitle{Embedded deformation for shape manipulation}.
\newblock In \bibinfo{booktitle}{\emph{ACM siggraph 2007 papers}}. \bibinfo{pages}{80--es}.
\newblock


\bibitem[Tagliasacchi et~al\mbox{.}(2012)]%
        {meso_skeleton}
\bibfield{author}{\bibinfo{person}{Andrea Tagliasacchi}, \bibinfo{person}{Ibraheem Alhashim}, \bibinfo{person}{Matt Olson}, {and} \bibinfo{person}{Hao Zhang}.} \bibinfo{year}{2012}\natexlab{}.
\newblock \showarticletitle{Mean Curvature Skeletons}.
\newblock \bibinfo{journal}{\emph{Computer Graphics Forum}} \bibinfo{volume}{31}, \bibinfo{number}{5} (\bibinfo{year}{2012}), \bibinfo{pages}{1735--1744}.
\newblock
\urldef\tempurl%
\url{https://doi.org/10.1111/j.1467-8659.2012.03178.x}
\showDOI{\tempurl}
\showeprint{https://onlinelibrary.wiley.com/doi/pdf/10.1111/j.1467-8659.2012.03178.x}


\bibitem[Tang et~al\mbox{.}(2022)]%
        {tang2022neural}
\bibfield{author}{\bibinfo{person}{Jiapeng Tang}, \bibinfo{person}{Lev Markhasin}, \bibinfo{person}{Bi Wang}, \bibinfo{person}{Justus Thies}, {and} \bibinfo{person}{Matthias Nie{\ss}ner}.} \bibinfo{year}{2022}\natexlab{}.
\newblock \showarticletitle{Neural shape deformation priors}.
\newblock \bibinfo{journal}{\emph{Advances in Neural Information Processing Systems}}  \bibinfo{volume}{35} (\bibinfo{year}{2022}), \bibinfo{pages}{17117--17132}.
\newblock


\bibitem[Terzopoulos et~al\mbox{.}(1987)]%
        {terzopoulos1987elastically}
\bibfield{author}{\bibinfo{person}{Demetri Terzopoulos}, \bibinfo{person}{John Platt}, \bibinfo{person}{Alan Barr}, {and} \bibinfo{person}{Kurt Fleischer}.} \bibinfo{year}{1987}\natexlab{}.
\newblock \showarticletitle{Elastically deformable models}. In \bibinfo{booktitle}{\emph{Proceedings of the 14th annual conference on Computer graphics and interactive techniques}}. \bibinfo{pages}{205--214}.
\newblock


\bibitem[Wang et~al\mbox{.}(2019)]%
        {wang20193dn}
\bibfield{author}{\bibinfo{person}{Weiyue Wang}, \bibinfo{person}{Duygu Ceylan}, \bibinfo{person}{Radomir Mech}, {and} \bibinfo{person}{Ulrich Neumann}.} \bibinfo{year}{2019}\natexlab{}.
\newblock \showarticletitle{3dn: 3d deformation network}. In \bibinfo{booktitle}{\emph{Proceedings of the IEEE/CVF Conference on Computer Vision and Pattern Recognition}}. \bibinfo{pages}{1038--1046}.
\newblock


\bibitem[Wen et~al\mbox{.}(2022)]%
        {wen20223d}
\bibfield{author}{\bibinfo{person}{Xin Wen}, \bibinfo{person}{Junsheng Zhou}, \bibinfo{person}{Yu-Shen Liu}, \bibinfo{person}{Hua Su}, \bibinfo{person}{Zhen Dong}, {and} \bibinfo{person}{Zhizhong Han}.} \bibinfo{year}{2022}\natexlab{}.
\newblock \showarticletitle{3D shape reconstruction from 2D images with disentangled attribute flow}. In \bibinfo{booktitle}{\emph{Proceedings of the IEEE/CVF conference on computer vision and pattern recognition}}. \bibinfo{pages}{3803--3813}.
\newblock


\bibitem[Xu et~al\mbox{.}(2020)]%
        {xu2020rignet}
\bibfield{author}{\bibinfo{person}{Zhan Xu}, \bibinfo{person}{Yang Zhou}, \bibinfo{person}{Evangelos Kalogerakis}, \bibinfo{person}{Chris Landreth}, {and} \bibinfo{person}{Karan Singh}.} \bibinfo{year}{2020}\natexlab{}.
\newblock \showarticletitle{RigNet: neural rigging for articulated characters}.
\newblock \bibinfo{journal}{\emph{ACM Transactions on Graphics (TOG)}} \bibinfo{volume}{39}, \bibinfo{number}{4} (\bibinfo{year}{2020}), \bibinfo{pages}{58--1}.
\newblock


\bibitem[Yang et~al\mbox{.}(2021)]%
        {yang2021NFGP}
\bibfield{author}{\bibinfo{person}{Guandao Yang}, \bibinfo{person}{Serge Belongie}, \bibinfo{person}{Bharath Hariharan}, {and} \bibinfo{person}{Vladlen Koltun}.} \bibinfo{year}{2021}\natexlab{}.
\newblock \showarticletitle{Geometry Processing with Neural Fields}. In \bibinfo{booktitle}{\emph{Thirty-Fifth Conference on Neural Information Processing Systems}}.
\newblock


\bibitem[Yifan et~al\mbox{.}(2020)]%
        {yifan2020neural}
\bibfield{author}{\bibinfo{person}{Wang Yifan}, \bibinfo{person}{Noam Aigerman}, \bibinfo{person}{Vladimir~G Kim}, \bibinfo{person}{Siddhartha Chaudhuri}, {and} \bibinfo{person}{Olga Sorkine-Hornung}.} \bibinfo{year}{2020}\natexlab{}.
\newblock \showarticletitle{Neural cages for detail-preserving 3d deformations}. In \bibinfo{booktitle}{\emph{Proceedings of the IEEE/CVF Conference on Computer Vision and Pattern Recognition}}. \bibinfo{pages}{75--83}.
\newblock


\bibitem[Yin et~al\mbox{.}(2020)]%
        {yin2020coalesce}
\bibfield{author}{\bibinfo{person}{Kangxue Yin}, \bibinfo{person}{Zhiqin Chen}, \bibinfo{person}{Siddhartha Chaudhuri}, \bibinfo{person}{Matthew Fisher}, \bibinfo{person}{Vladimir~G Kim}, {and} \bibinfo{person}{Hao Zhang}.} \bibinfo{year}{2020}\natexlab{}.
\newblock \showarticletitle{Coalesce: Component assembly by learning to synthesize connections}. In \bibinfo{booktitle}{\emph{2020 International Conference on 3D Vision (3DV)}}. IEEE, \bibinfo{pages}{61--70}.
\newblock


\bibitem[Yin et~al\mbox{.}(2014)]%
        {Morfit2014}
\bibfield{author}{\bibinfo{person}{Kangxue Yin}, \bibinfo{person}{Hui Huang}, \bibinfo{person}{Hao Zhang}, \bibinfo{person}{Minglun Gong}, \bibinfo{person}{Daniel Cohen-Or}, {and} \bibinfo{person}{Baoquan Chen}.} \bibinfo{year}{2014}\natexlab{}.
\newblock \showarticletitle{Morfit: interactive surface reconstruction from incomplete point clouds with curve-driven topology and geometry control}.
\newblock \bibinfo{journal}{\emph{ACM Trans. Graph.}} \bibinfo{volume}{33}, \bibinfo{number}{6}, Article \bibinfo{articleno}{202} (\bibinfo{date}{nov} \bibinfo{year}{2014}), \bibinfo{numpages}{12}~pages.
\newblock
\showISSN{0730-0301}
\urldef\tempurl%
\url{https://doi.org/10.1145/2661229.2661241}
\showDOI{\tempurl}


\bibitem[Yoshizawa et~al\mbox{.}(2007)]%
        {yoshizawa2007skeleton}
\bibfield{author}{\bibinfo{person}{Shin Yoshizawa}, \bibinfo{person}{Alexander Belyaev}, {and} \bibinfo{person}{Hans-Peter Seidel}.} \bibinfo{year}{2007}\natexlab{}.
\newblock \showarticletitle{Skeleton-based variational mesh deformations}. In \bibinfo{booktitle}{\emph{Computer Graphics Forum}}, Vol.~\bibinfo{volume}{26}. Wiley Online Library, \bibinfo{pages}{255--264}.
\newblock


\bibitem[Yoshizawa et~al\mbox{.}(2003)]%
        {yoshizawa2003free}
\bibfield{author}{\bibinfo{person}{Shin Yoshizawa}, \bibinfo{person}{Alexander~G Belyaev}, {and} \bibinfo{person}{Hans-Peter Seidel}.} \bibinfo{year}{2003}\natexlab{}.
\newblock \showarticletitle{Free-form skeleton-driven mesh deformations}. In \bibinfo{booktitle}{\emph{Proceedings of the eighth ACM symposium on Solid modeling and applications}}. \bibinfo{pages}{247--253}.
\newblock


\bibitem[Yu et~al\mbox{.}(2011)]%
        {curve_length_cons}
\bibfield{author}{\bibinfo{person}{R. Yu}, \bibinfo{person}{R. Wang}, {and} \bibinfo{person}{Chun-Gang Zhu}.} \bibinfo{year}{2011}\natexlab{}.
\newblock \showarticletitle{Curve interpolation with length constraint in a discrete manner}.
\newblock \bibinfo{journal}{\emph{Journal of Information and Computational Science}}  \bibinfo{volume}{8} (\bibinfo{date}{06} \bibinfo{year}{2011}), \bibinfo{pages}{859--868}.
\newblock


\bibitem[Zhang et~al\mbox{.}(2023)]%
        {3DShape2VecSet}
\bibfield{author}{\bibinfo{person}{Biao Zhang}, \bibinfo{person}{Jiapeng Tang}, \bibinfo{person}{Matthias Nie\ss{}ner}, {and} \bibinfo{person}{Peter Wonka}.} \bibinfo{year}{2023}\natexlab{}.
\newblock \showarticletitle{3DShape2VecSet: A 3D Shape Representation for Neural Fields and Generative Diffusion Models}.
\newblock \bibinfo{journal}{\emph{ACM Trans. Graph.}} \bibinfo{volume}{42}, \bibinfo{number}{4}, Article \bibinfo{articleno}{92} (\bibinfo{date}{jul} \bibinfo{year}{2023}), \bibinfo{numpages}{16}~pages.
\newblock
\showISSN{0730-0301}
\urldef\tempurl%
\url{https://doi.org/10.1145/3592442}
\showDOI{\tempurl}


\bibitem[Zheng et~al\mbox{.}(2011)]%
        {Zheng2011Component-wise-Controllers}
\bibfield{author}{\bibinfo{person}{Youyi Zheng}, \bibinfo{person}{Hongbo Fu}, \bibinfo{person}{Daniel Cohen-Or}, \bibinfo{person}{Oscar Kin-Chung Au}, {and} \bibinfo{person}{Chiew-Lan Tai}.} \bibinfo{year}{2011}\natexlab{}.
\newblock \showarticletitle{Component-wise Controllers for Structure-Preserving Shape Manipulation}.
\newblock \bibinfo{journal}{\emph{Computer Graphics Forum}} \bibinfo{volume}{30}, \bibinfo{number}{2} (\bibinfo{year}{2011}), \bibinfo{pages}{563--572}.
\newblock
\urldef\tempurl%
\url{https://doi.org/10.1111/j.1467-8659.2011.01880.x}
\showDOI{\tempurl}
\showeprint{https://onlinelibrary.wiley.com/doi/pdf/10.1111/j.1467-8659.2011.01880.x}


\bibitem[Zhou et~al\mbox{.}(2005)]%
        {zhou2005large}
\bibfield{author}{\bibinfo{person}{Kun Zhou}, \bibinfo{person}{Jin Huang}, \bibinfo{person}{John Snyder}, \bibinfo{person}{Xinguo Liu}, \bibinfo{person}{Hujun Bao}, \bibinfo{person}{Baining Guo}, {and} \bibinfo{person}{Heung-Yeung Shum}.} \bibinfo{year}{2005}\natexlab{}.
\newblock \showarticletitle{Large mesh deformation using the volumetric graph laplacian}.
\newblock In \bibinfo{booktitle}{\emph{ACM SIGGRAPH 2005 Papers}}. \bibinfo{pages}{496--503}.
\newblock


\bibitem[Zhou et~al\mbox{.}(2015)]%
        {gcd15}
\bibfield{author}{\bibinfo{person}{Yang Zhou}, \bibinfo{person}{Kangxue Yin}, \bibinfo{person}{Hui Huang}, \bibinfo{person}{Hao Zhang}, \bibinfo{person}{Minglun Gong}, {and} \bibinfo{person}{Daniel Cohen-Or}.} \bibinfo{year}{2015}\natexlab{}.
\newblock \showarticletitle{Generalized Cylinder Decomposition}.
\newblock \bibinfo{journal}{\emph{ACM Transactions on Graphics (Proc. of SIGGRAPH Asia)}} \bibinfo{volume}{34}, \bibinfo{number}{6} (\bibinfo{year}{2015}), \bibinfo{pages}{171:1--171:14}.
\newblock


\end{thebibliography}


\end{document}